\documentclass{tADR2e}

\usepackage{xcolor}
\usepackage{gensymb}

\begin{document}

\jvol{00} \jnum{00} \jyear{2021} \jmonth{January}

\articletype{Full paper}

\title{Selecting and Designing Grippers for an Assembly Task \\in a Structured Approach}

\author{Jingren Xu$^{a,b}$$^{\ast}$\thanks{$^\ast$Corresponding author. Email: xu@hlab.sys.es.osaka-u.ac.jp
\vspace{6pt}}, Weiwei Wan$^a$, Keisuke Koyama$^a$, Yukiyasu Domae$^b$, and Kensuke Harada$^{a,b}$ \\\vspace{6pt}  $^{a}${Graduate School of Engineering Science, Osaka University, Osaka, Japan};\\
$^{b}${National Institute of Advanced Industrial Science and Technology (AIST), Tokyo, Japan}\\\vspace{6pt}\received{v1.0 released January 2021} }

\maketitle

\begin{abstract}
In this paper, we present a structured approach to selecting and designing a set of grippers for an assembly task. Compared to current experience-based gripper design method, our approach accelerates the design process by automatically generating a set of initial design options for gripper types and parameters according to the CAD models of assembly components. We use mesh segmentation techniques to segment the assembly components and fit the segmented parts with shape primitives, according to the predefined correspondence between shape primitive and gripper type, suitable gripper types and parameters can be selected and extracted from the fitted shape primitives. Moreover, we incorporate the assembly constraints in the further evaluation of the initially obtained gripper types and parameters. Considering the affordance of the segmented parts and the collision avoidance between the gripper and the subassemblies, applicable gripper types and parameters can be filtered from the initial options. Among the applicable gripper configurations, we further optimize the number of grippers for performing the assembly task, by exploring the gripper that is able to handle multiple assembly components during the assembly. Finally, the feasibility of the designed grippers is experimentally verified by assembling a part of an industrial product.

\medskip

\end{abstract}

\section{Introduction}
Robots have been increasing engaged in industrial applications such as robotic assembly, where a set of mechanical components are handled and manipulated by robotic grippers. The gripper plays a pivotal role for the robot interacting with the environment, the performance of the gripper grasping an assembly component is strongly influenced by how well the chosen gripper and its characteristics coincide with the characteristics needed for grasping a specific part \cite{schmalz2014automated}. Therefore, designing reliable grippers is one of the key issues for applying robots in industrial environment.

However, robotic grippers are manually designed in most of the cases, the manual design process is time-consuming and requires a lot of experience and expertise, which makes it extremely challenging to design grippers, especially for an assembly task. In a general robotic assembly task, a set of specialized grippers are required to firmly grasp all the assembly components with different shapes and properties, in addition, the grippers have to satisfy the assembly constraints, such as avoiding collision with other subassemblies. Moreover, there is a trend in High-Mix Low-Volume production, which refers to producing a large variety of products in small quantities, the fast changing manufacturing routines propose great challenges for applying robot in such agile manufacturing. Therefore, in terms of the grippers used in the assembly tasks, a more efficient approach of designing grippers is highly demanded in order to quickly adapt to the frequently changing assembly tasks.

\begin{figure}
    \centering
    \includegraphics[width=0.65\textwidth]{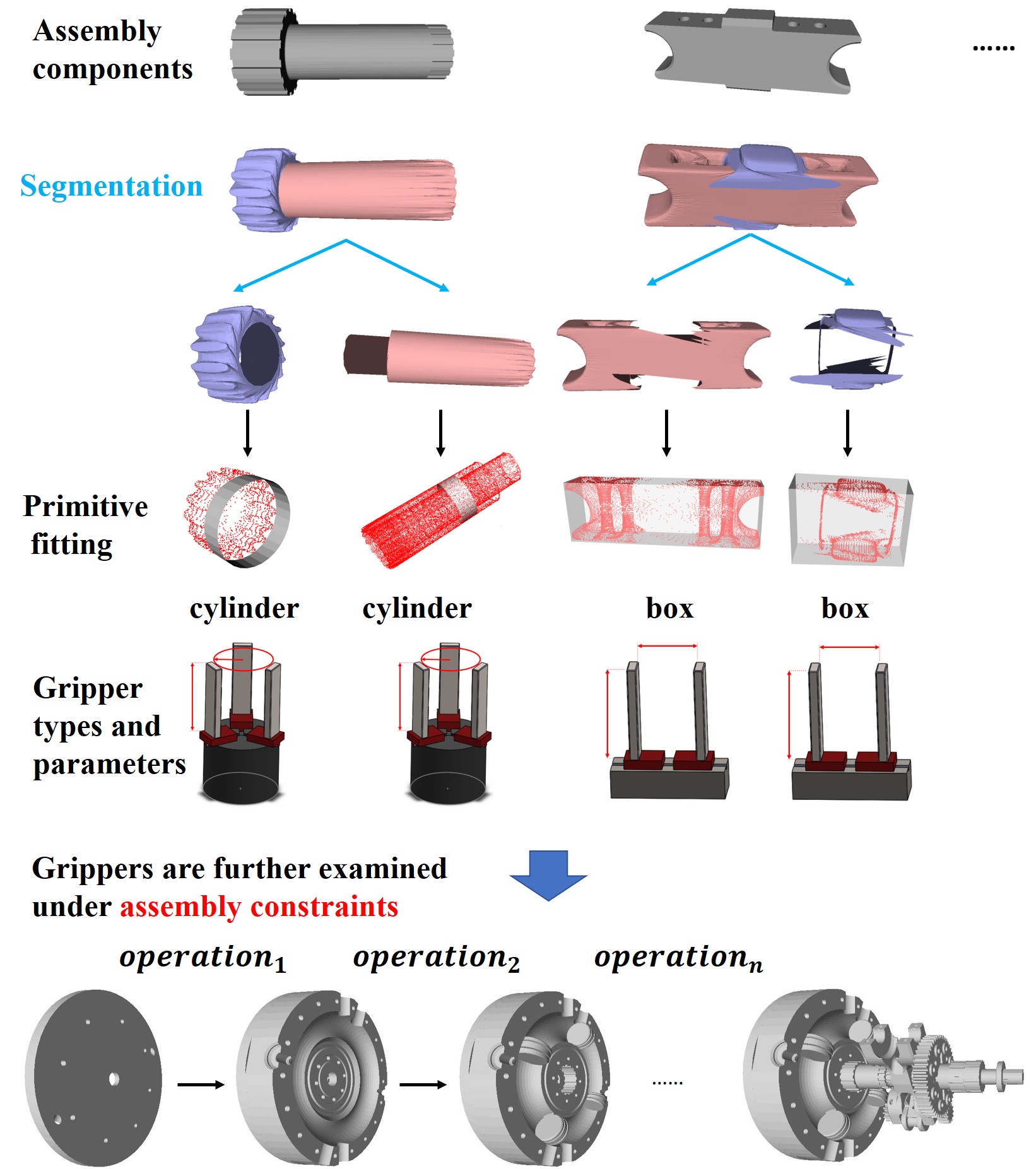}
    \caption{Overview of the proposed approach of selecting and designing grippers for an assembly task. In the first stage, suitable gripper types (2-finger or 3-finger gripper) and parameters (opening width) can determined by mesh segmentation and primitive fitting. Then the segments and grippers of such configurations are further evaluated under the assembly constraints, such as affordance and collision avoidance. Finally, we optimize the number of grippers to cut down the total cost.}
    \label{fig:overview}
\end{figure}

To efficiently design grippers satisfying the assembly constraints, we propose a structured approach of selecting and designing the grippers based on the shape analysis and assembly constraints, the overview of the method is illustrated in Fig. \ref{fig:overview}. The insight is that the industrial products are usually comprised of many regular shape primitives, such as cylinder and cuboid, each of the shape primitives can be firmly grasped by a suitable type of gripper. Therefore, we pre-define the rules for selecting suitable gripper types, which reduces the space for searching possible gripper configurations and significantly accelerates the design process. Through mesh segmentation techniques, we can uncover the underlying shape primitives and assign predefined gripper types to them. The gripper parameters, such as the maximum and minimum opening widths, can be further extracted from the dimensions of the fitted primitives. These steps are automatically processed and provide reduced gripper configurations for further selection and evaluation. These gripper configurations work well in terms of grasping, however, robotic assembly is a much more complex task, where the grippers have to not only firmly grasp the assembly components, but also avoid the collision with the subassemblies. Furthermore, some segments are not suitable for grasping considering their affordance, and they are excluded from the selection of graspable segments. After the evaluation under assembly constraints, some of the remaining segments can be commonly grasped by the same gripper, therefore, the number of grippers can be optimized to reduce the total cost.

A few researches were performed on the design of grippers for an assembly task. However these researches are limited to designing the local shape of the fingertip \cite{honarpardaz2019fast} and general suggestions for designing the gripper systems \cite{pham1991strategies}. There has been no attempt on the structured approach of selecting and designing grippers according to the assembly constraints, as well as minimizing the number of grippers, for an assembly task.

The rest of the paper is organized as follows: Section 2 reviews the related work. Section 3 introduces the segmentation of the assembly components. In section 4, the initial set of gripper configurations is extracted by primitive fitting. In section 5, we evaluate these gripper configurations under the assembly constraints and optimize the number of grippers. The feasibility of the designed grippers are confirmed by an assembly experiment in section 6. Finally, we draw the conclusions and provide the prospect for future work in section 7.

\section{Related Work}
There has been a lot of research on gripper design \cite{cannon2005compliant,nie2018adaptive}, however, very few of them designs the grippers for an assembly task considering the assembly constraints. Another line of research that is related to our work is part/model/primitive-based grasp planning \cite{miller2003automatic,przybylski2011planning,huebner2008selection,vahrenkamp2016part,bohg2013data}, in this sense, our work can be called shape primitive-based gripper design, considering assembly constraints and optimization of number of grippers.

\subsection{Gripper Design and Robotic Assembly}
Generally, the grippers are specially designed according to the task to be performed \cite{nie2018adaptive, cannon2005compliant}, in this case, the design process takes many iterations to obtain a satisfactory design. There have been very few attempts to design grippers for an assembly task and improve the design efficiency, to the best of our knowledge, the most relevant works to ours are \cite{pham1991strategies, pham2007automated}. Pham et al. \cite{pham1991strategies} surveyed the design methods to achieve versatile and cost-effective gripping and proposed a strategy for minimizing the number of grippers through part-family grouping, and later Pham et al. \cite{pham2007automated} proposed a system to determine the configuration of grippers for an assembly task. However, none of the these works explicitly incorporate the assembly constraints into the gripper design, besides, the mesh segmentation and primitive fitting method used in our approach is able to handle models with more complex shapes.

In addition to the gripper configuration, the contact between the gripper finger and object plays an important role on grasp stability, therefore, the contact model has been studied extensively \cite{harada2014stability,ciocarlie2007soft,harada2011grasp,bicchi2000robotic}. Early research mainly used point contact model \cite{murray2017mathematical}, later on, soft finger model was developed to model the contact in a more realistic way \cite{ciocarlie2007soft,harada2014stability}. Some researchers studied the finger design to change the contact characteristics and improve the performance of the gripper. Honarpardaz et al. \cite{honarpardaz2017fast,honarpardaz2019fast} proposed generic optimized finger design (GOFD) to automate the finger design process, the fingertip shape was designed to mimic the local surface contour of workpiece, thus the contact area was increased. Song et al. \cite{song2018fingertip} noticed that most grasp contacts share a few local geometries, they proposed a uniform cost algorithm to cluster a set of example grasp contacts into several contact primitives, and designed the fingertip shape to match the local geometry of the contact primitive in order to increase the contact area.

Rodriguez et al. \cite{rodriguez2013effector} explored the effector form design for 1 DoF planar actuation, the mechanical function of a product is formulated as the product of the effector's shape and motion. Taylor el al. \cite{taylor2019optimal} investigated the role of shape and motion in the contact interaction, and proposed a framework to optimize the shape and motion of a planar rigid body end-effector to achieve a manipulation task. Chavan-Dafle et al. \cite{chavan2015two} proposed a two-phase gripper to passively reorient the objects while picking them up. Birglen et al. \cite{birglen2018statistical} extensively reviewed the characteristics of industrial grippers, the stroke, weight, force and weight, as well as performance, are investigated in detail. Hermann et al. \cite{hermann2019joint} designed a gripper that can switch between two modes, including a grasping mode and a fully actuated precision mode.

For an assembly task, usually more than one gripper is required to grasp all the assembly components. Kramberger et al. \cite{kramberger2019automatic} proposed a flexible and cost-effective grasping solution to quickly develop and test fingertips to handle multiple parts. Harada et al. \cite{harada2018tool} incorporated the tool changer into the assembly planner and proposed an assembly planner that is able to automatically select a suitable gripper to assemble parts. Nakayama et al. \cite{nakayama2019designing} designed grasping tools for an assembly task based on shape analysis of parts, however, the assembly constraints are not considered in the evaluation of graspable segments and suitable gripper configurations, and additionally we optimize the number of grippers for the assembly task.

\subsection{Shape Approximation Based Grasping}
Grasp planning is difficult due to the large number of possible gripper configurations, but grasping planning can be simplified if considering the shape of the object and the grasping strategy are closely related. Miller et al. modeled the object as a set of simple shape primitives \cite{miller2003automatic}, then the grasp location and preshape can be determined. Goldfeder et. al \cite{goldfeder2007grasp} used a decomposition tree of the object to prune the large space of possible grasps into a subspace that is likely to contain many good grasps. Huebner et al. \cite{huebner2008selection} approximated the object by box primitives and selected grasps based on the approximated boxes. However, the error of approximation by primitives may result in low-quality grasps, to counteract this problem, Przybylski et al. \cite{przybylski2011planning} proposed the grid of spheres for grasp planning, which effectively reduces the search space for grasps without sacrificing potential high-quality grasps.

These researches \textbf{passively} plan grasps given the object model, but we can also \textbf{actively} design the gripper configurations according the shape of the target object in order to easily obtain high-quality grasps. This idea is somewhat related to the taxonomy of grasps proposed in \cite{cutkosky1989grasp}, where the grasps are classified based on task-related and geometric considerations, each type of grasps is corresponding to one category of tasks and object geometry. For grasping the assembly components, we select suitable grasping postures according to the shape of the assembly components, since we do not use dexterous robot hand to realize these grasps, instead we abstract a simple gripper configuration from the grasping postures of dexterous hand.

Our main contributions are summarized as follows:
\begin{itemize}
    \item We proposed a structured approach of configuring the gripper types and parameters based on mesh segmentation, primitive fitting and assembly constraints.
    \item The assembly constraints are explicitly taken into account in the evaluation of the feasible gripper configurations. The number of grippers required for the assembly task is optimized to reduce the cost.
    \item The proposed method is experimentally verified by assembling a part of an industrial product.
\end{itemize}

\section{Mesh Segmentation}

\begin{figure}
    \centering
    \includegraphics[width=0.7\textwidth]{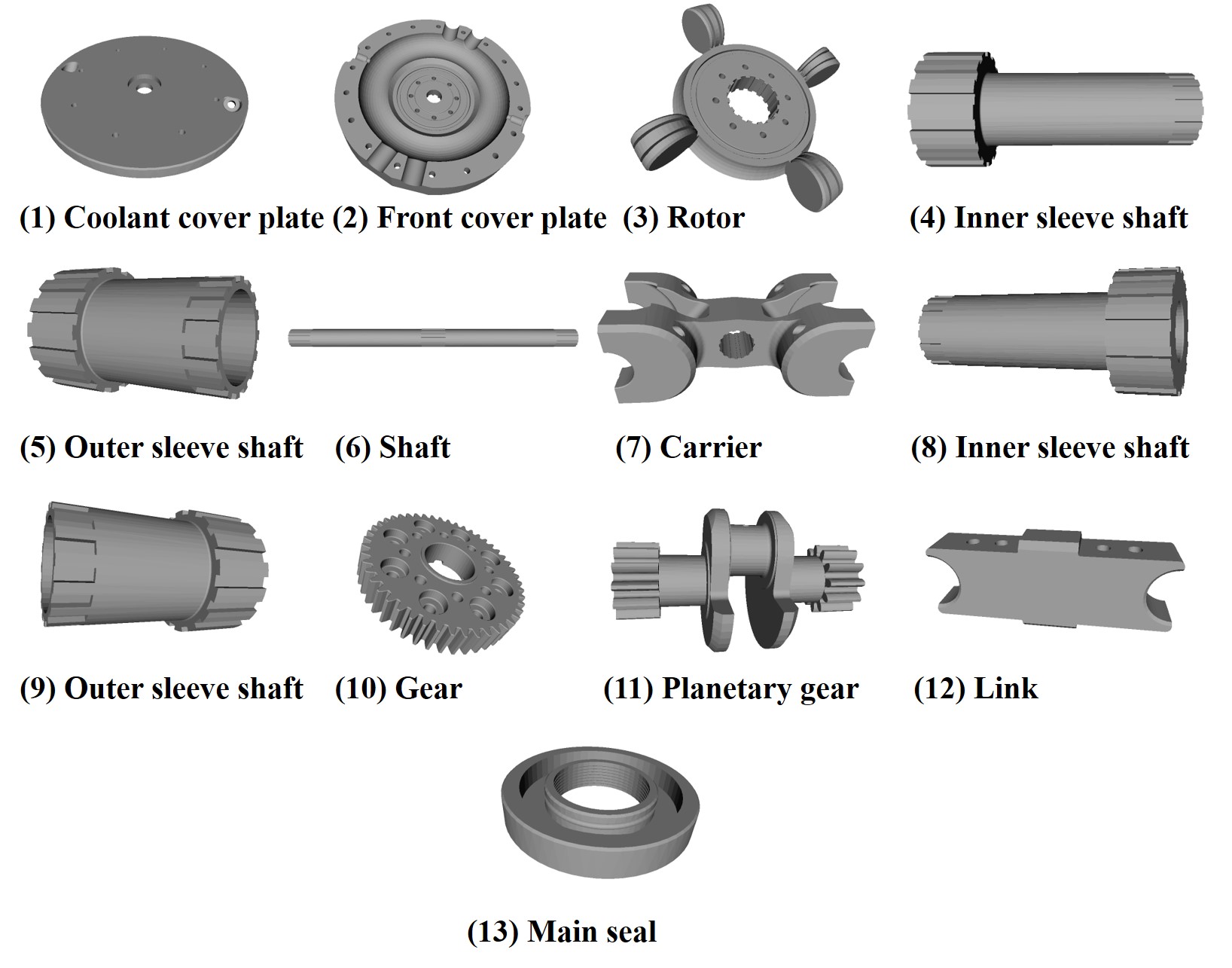}
    \caption{Models of all the assembly components before processing, they are displayed in the order of the assembly sequence.}
    \label{fig:all_components_before_processing}
\end{figure}

\begin{figure}
\centering
    \subfigure[]{\includegraphics[width=0.6\textwidth]{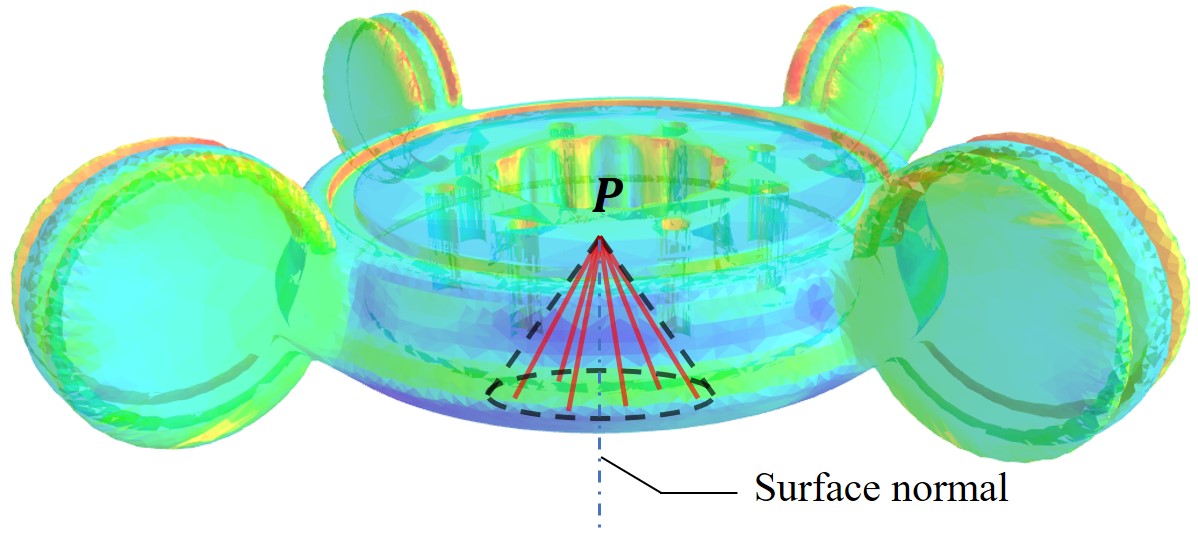}}
    \subfigure[]{\includegraphics[width=0.4\textwidth]{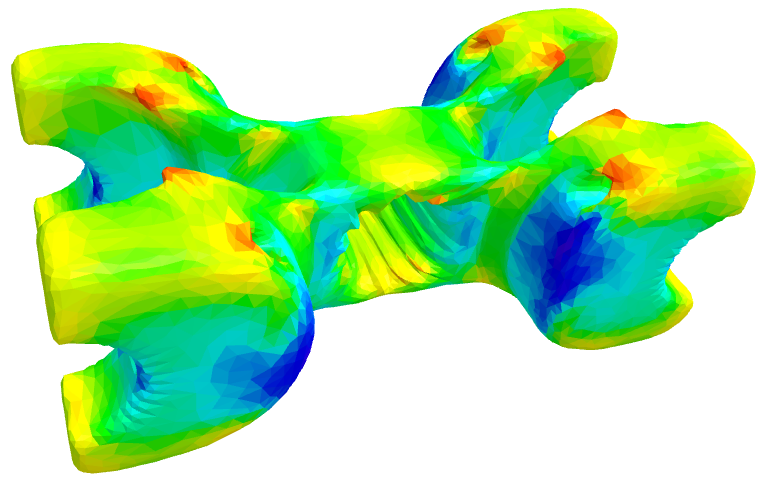}}
    \subfigure[]{\includegraphics[width=0.4\textwidth]{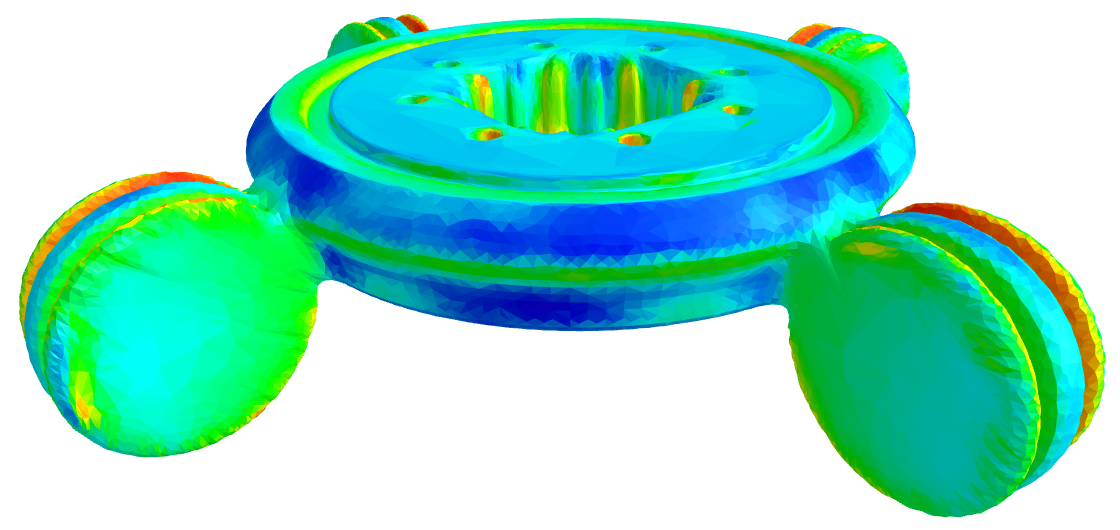}}
    \caption{(a) The Shape Diameter Function (SDF) is the weighted average length of the rays (red lines). (b) \& (c) SDF distribution of the carrier and the rotor.}
    \label{fig:sdf}
\end{figure}

Mechanical products are usually comprised of many regular shapes such as cylinders and cuboids, which makes the proposed method feasible and promising in industrial applications. We use mesh segmentation to find the underlying shape primitives of an assembly component, then suitable gripper types are determined according to the predefined rules. The mesh models of the assembly components (Fig. \ref{fig:all_components_before_processing}) are segmented based on Shape Diameter Function (SDF) \cite{shapira2008consistent}, which is a scalar function measuring the neighborhood diameter of an object at each point on the surface. To obtain the SDF value at a point $P$ on the surface, we construct a cone centered around the inward-normal direction of $P$, as sketched in black dashed lines in Fig. \ref{fig:sdf} (a), from $P$ we shoot a set of rays (red lines) inside the cone and stop at the intersections on another side of the mesh. The SDF value is calculated as the weighted average length of the rays. In our implementation, we shoot 30 rays per point and set the cone angle to $120 \degree$, as a result, Fig. \ref{fig:sdf} (b) \& (c) show two examples of SDF distribution on the model. The mesh segmentation process is comprised of soft clustering and hard clustering. Soft clustering is a Gaussian mixture model that fits a set of Gaussian distributions to the distribution of the SDF values, this step outputs the probability matrix for each face to belong to each cluster, note that a cluster may contain multiple segments. Hard clustering yields the final segmentation of the mesh by minimizing an energy function combining the probability matrix and geometric surface features \cite{shapira2008consistent,cgal:eb-20a}. Readers are referred to \cite{shamir2008survey} for other mesh segmentation methods.

\begin{figure}
    \centering
    \includegraphics[width=0.5\textwidth]{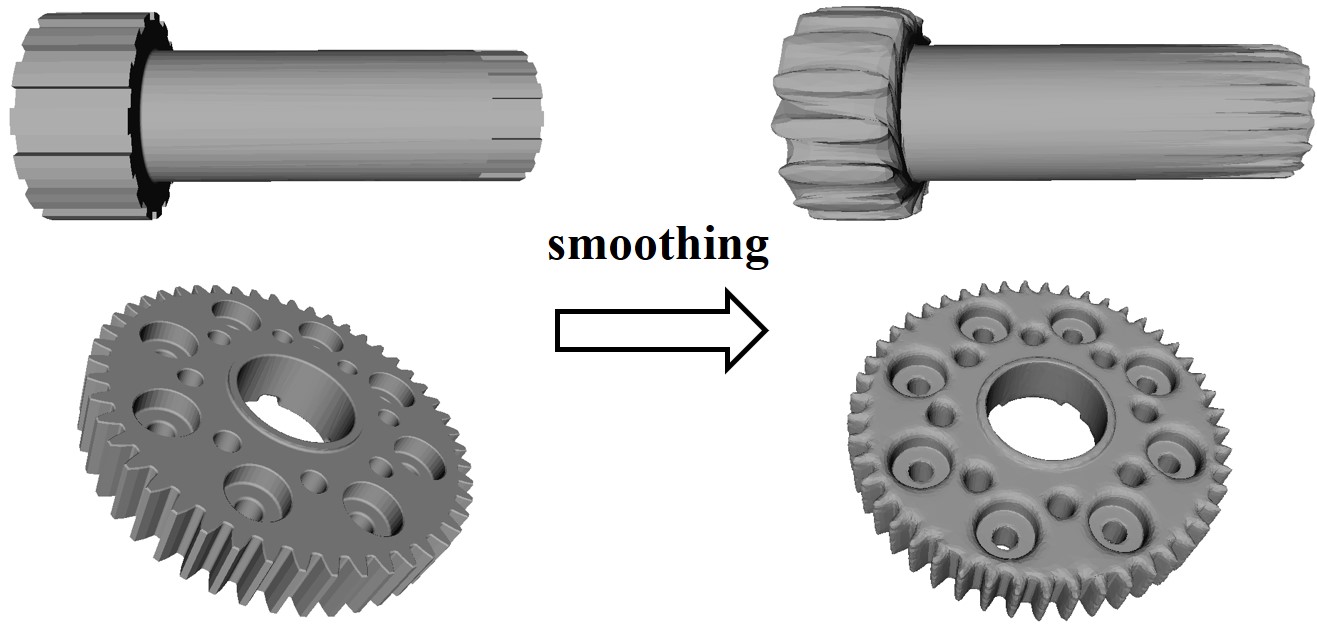}
    \caption{Two examples of models before and after smoothing.}
    \label{fig:smoothing}
\end{figure}

\begin{figure}
\centering
    \subfigure{\includegraphics[width=0.19\linewidth]{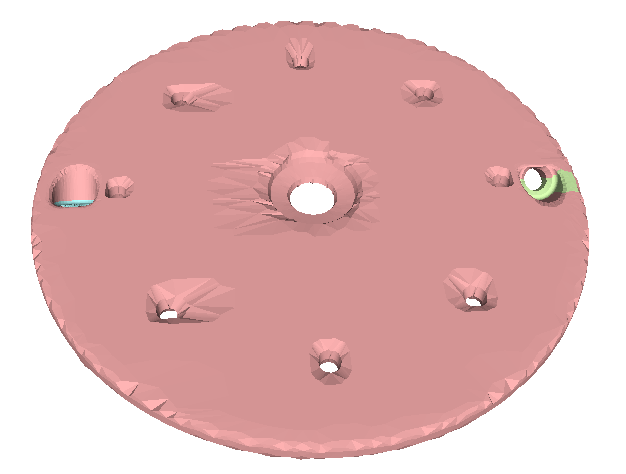}}
    \subfigure{\includegraphics[width=0.19\linewidth]{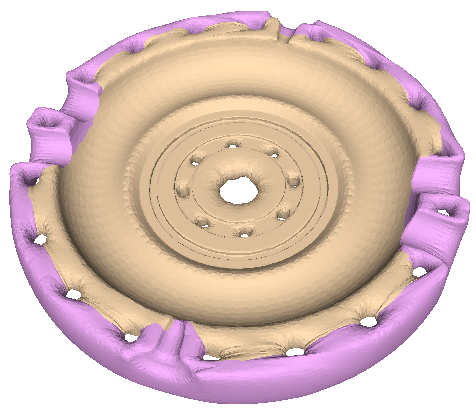}}
    \subfigure{\includegraphics[width=0.19\linewidth]{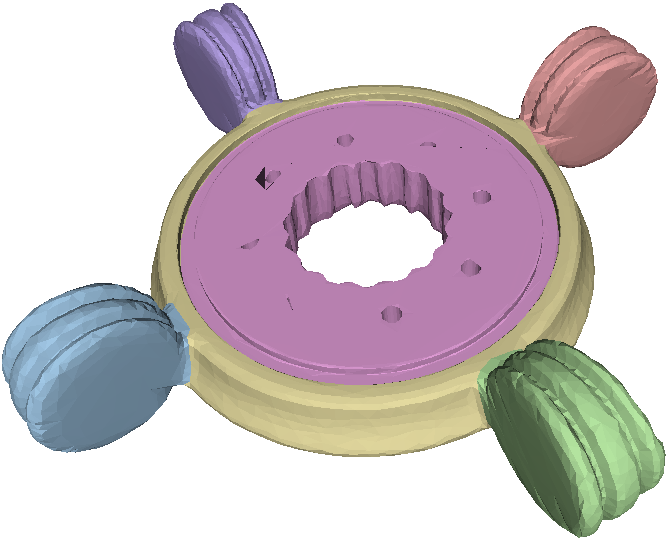}}
    \subfigure{\includegraphics[width=0.19\linewidth]{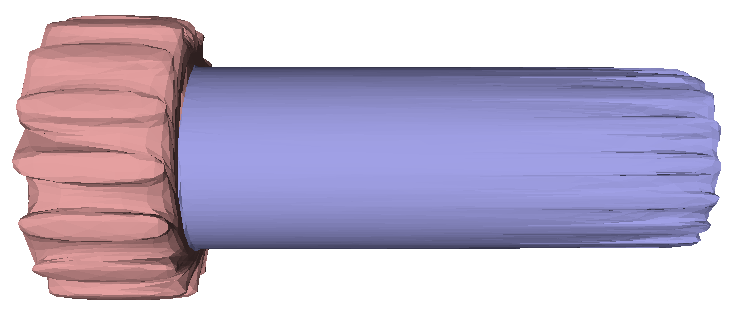}}
    \subfigure{\includegraphics[width=0.19\linewidth]{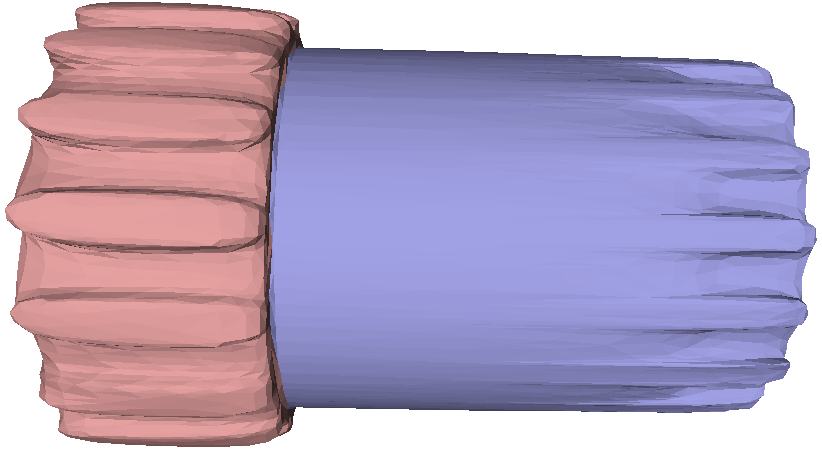}}
    \subfigure{\includegraphics[width=0.19\linewidth]{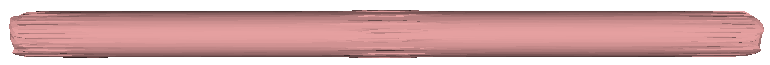}}
    \subfigure{\includegraphics[width=0.19\linewidth]{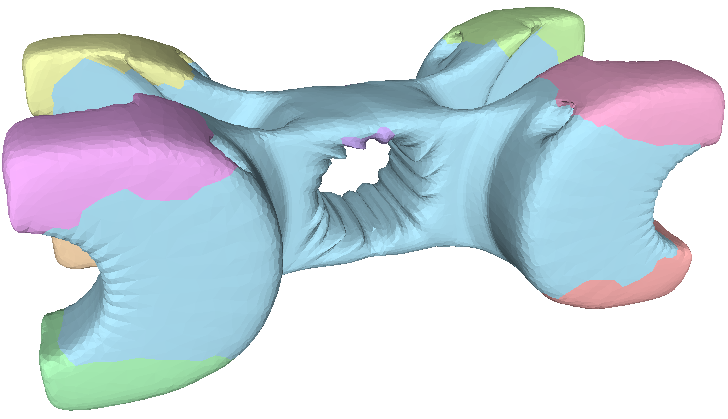}}
    \subfigure{\includegraphics[width=0.19\linewidth]{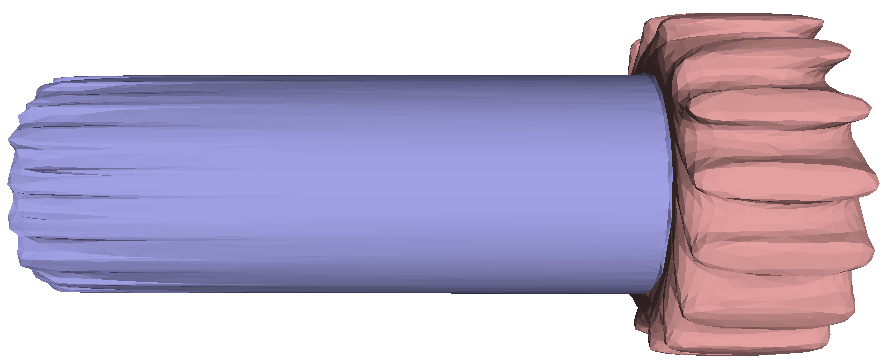}}
    \subfigure{\includegraphics[width=0.19\linewidth]{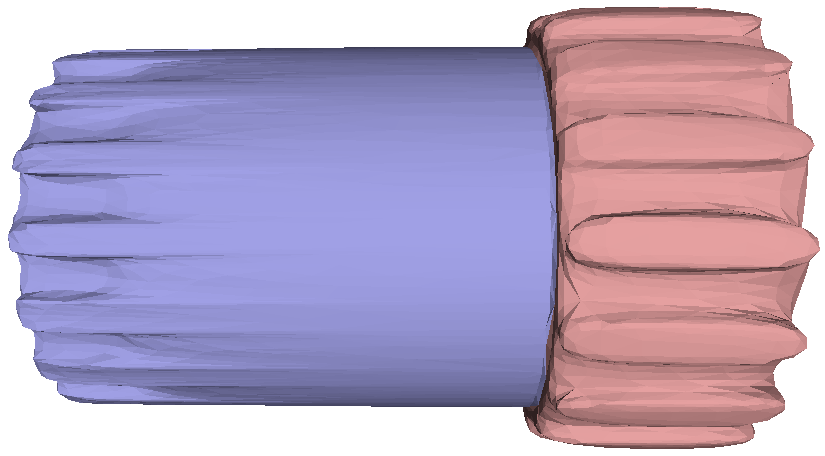}}
    \subfigure{\includegraphics[width=0.19\linewidth]{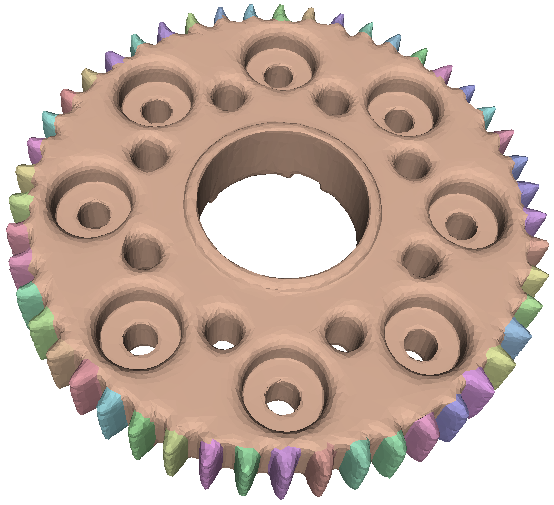}}
    \subfigure{\includegraphics[width=0.19\linewidth]{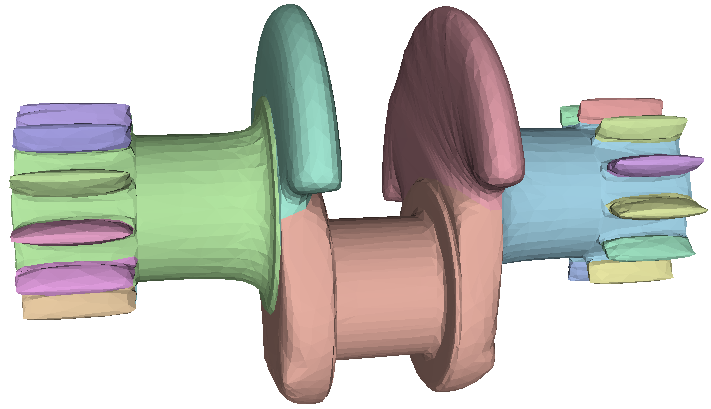}}
    \subfigure{\includegraphics[width=0.19\linewidth]{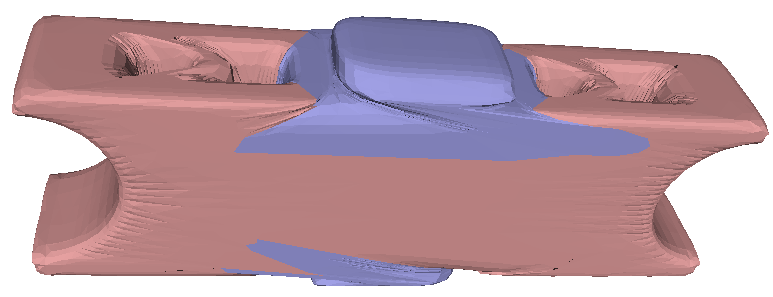}}
    \subfigure{\includegraphics[width=0.19\linewidth]{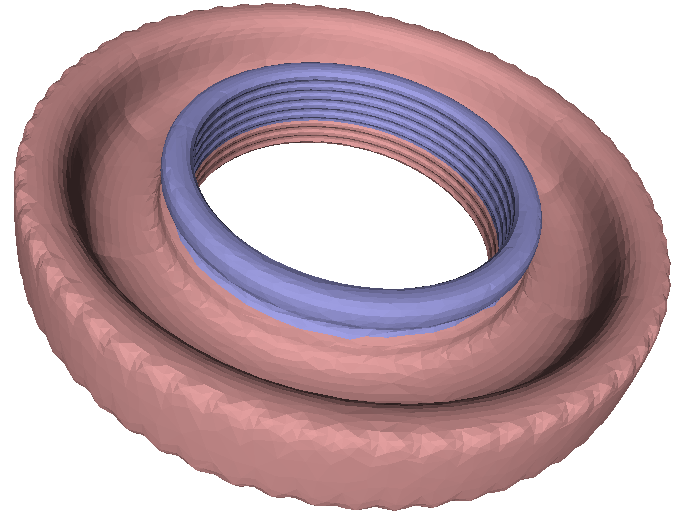}}
    \caption{After mesh segmentation, an assembly component is decomposed into several segments, different segments are rendered with different colors. The original component with a complex shape is decomposed into segments with simpler shapes, which are suitable for further primitive fitting.}
    \label{fig:segmentation}
\end{figure}

Before mesh segmentation, smoothing is applied on the mesh to eliminate the sharp edges of the screw thread, otherwise it may result in undesirable segments \cite{cignoni2008meshlab}. Fig. \ref{fig:smoothing} shows the mesh after smoothing is applied. Then all the assembly components are segmented based on SDF values. The segmentation results are visualized as Fig. \ref{fig:segmentation}, different segments are colored differently, each of the segments is regarded as a candidate for grasping\footnote{Simultaneously grasping multiple segments is not considered in this paper.}. 

\section{Gripper Selection and Dimensioning}
Through mesh segmentation, an assembly component with a complex shape is decomposed into segments with simpler shapes. Obviously, some shape primitives can be easily grasped by some common types of grippers, e.g. cylinders can be easily grasped by the 3-finger centric gripper. Therefore, we attempt to fit the segments with shape primitives and then determine the suitable gripper types according to the predefined rules. In this section, we obtain the initial decision on gripper types and parameters based on previous segmentation results.

\subsection{Rules for Gripper Type Selection}
\begin{figure}
    \centering
    \includegraphics[width=0.7\textwidth]{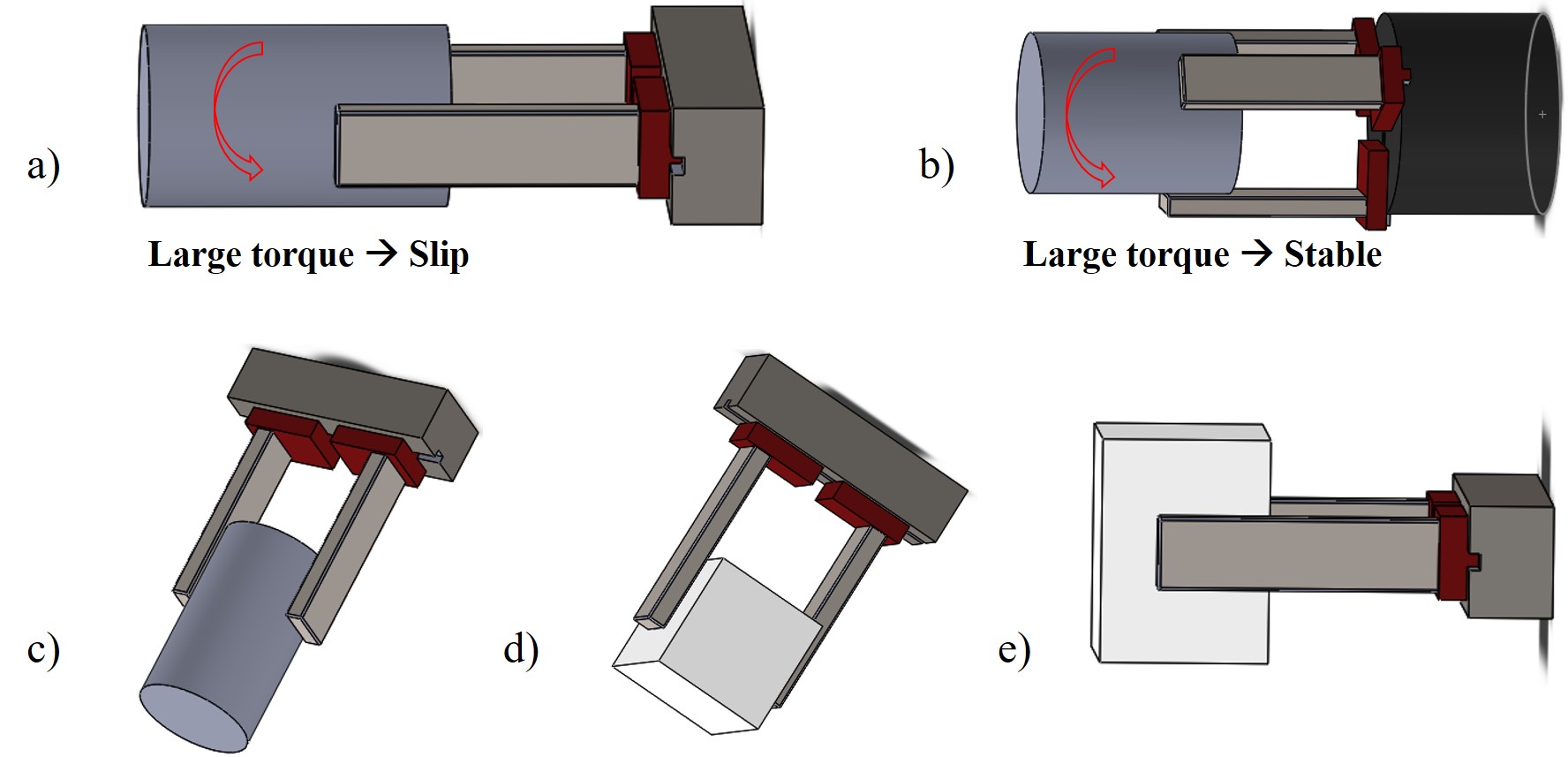}
    \caption{(a) \& (c) Grasping a cylinder by a 2-finger gripper is not stable against large external torques in cylinder's radial direction, the object may slip around the contact normal. (b) Grasping a cylinder by a 3-finger gripper is stable against large external torques in cylinder's radial direction. (d) \& (e) Grasping a box shape by a 2-finger gripper is appropriate.}
    \label{fig:gripper_rules}
\end{figure}

\begin{figure}
    \centering
    \includegraphics[width=0.5\textwidth]{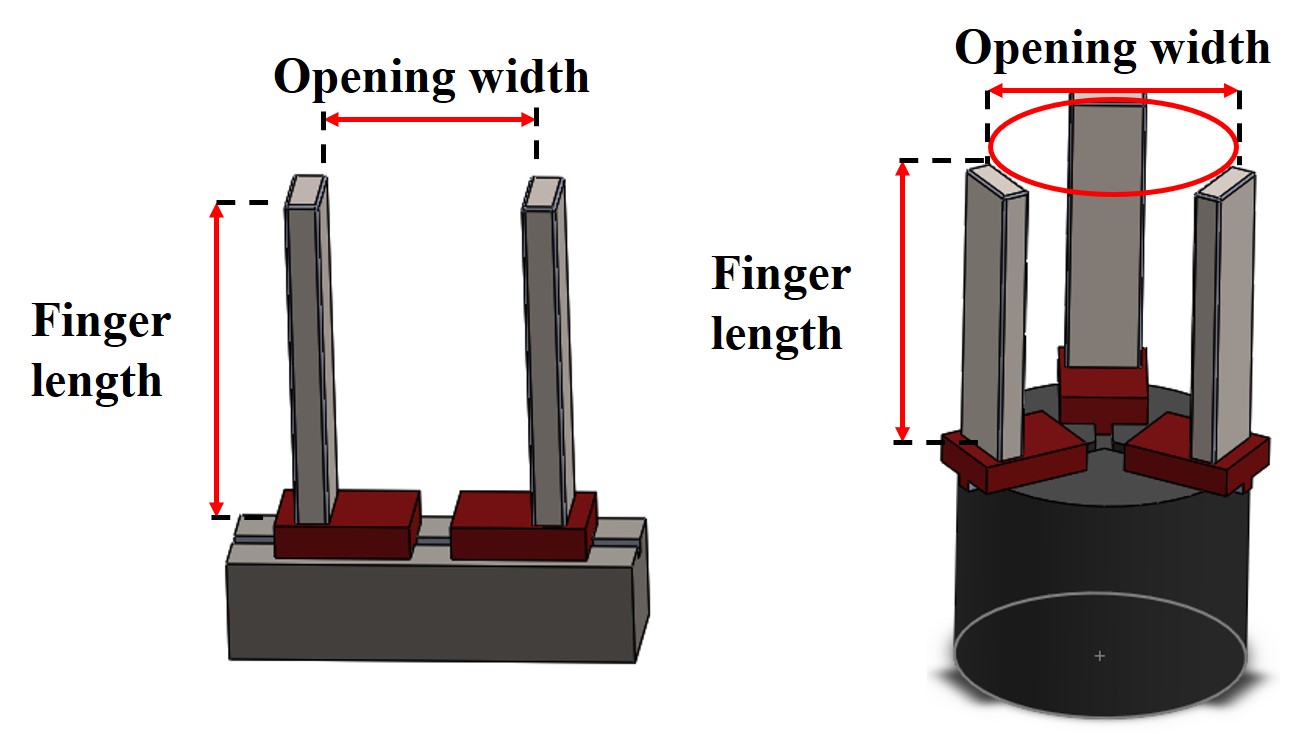}
    \caption{The opening width and finger length of the 2-finger and 3-finger grippers.}
    \label{fig:gripper_parameters}
\end{figure}

In this paper, we consider using two common types of grippers\footnote{More gripper types and shape primitives can be used to cope with more complex shapes.}: 2-finger parallel jaw grippers and 3-finger centric grippers as shown in Fig. \ref{fig:gripper_rules} (a) and (b). 2-finger grippers are suitable for grasping parts with (nearly) parallel surfaces, Fig. \ref{fig:gripper_rules} (d) and (e) show a 2-finger gripper grasping a box with parallel surfaces, the gripper fingers coincide well with the object surfaces and they have large contact area, thus the grasp is stable. However, it may not be suitable to grasp a cylinder using a 2-finger gripper, as shown in Fig. \ref{fig:gripper_rules} (a), small external torques in cylinder's radial direction can be balanced by assuming soft-finger contact\footnote{The object and the main body of the finger are assumed to be rigid, but soft pad can be attached to the fingertip.}, but in some assembly operations the gripper may have to exert large forces/torques on the assembly components, which may lead to slip around the contact normal. Therefore, we favor 3-finger centric grippers over 2-finger grippers for grasping cylindrical objects, which is guaranteed to be stable against the torque in the radial direction, as shown in Fig. \ref{fig:gripper_rules} (b). Another merit of grasping cylindrical objects using 3-finger grippers is that the grasp stability is independent of the radius of the cylinder, however, the stability of grasping a cylindrical object using a 2-finger gripper depends on the relative curvature of the finger and object surface, that is, grasping a cylinder with larger radius is more stable since the contact area is larger\footnote{Assume soft finger contact and constant external forces.}.


\subsection{Gripper Type}
Each segment of an assembly component shown in Fig. \ref{fig:segmentation} is a candidate for grasping, in order to determine a suitable gripper type for grasping the segment, we fit every segment with a cylinder and a bounding box. If the volume of cylinder is closer to the volume of the segment, then a 3-finger centric gripper is selected for this segment, otherwise, the 2-finger jaw gripper is used. Since the segments of a surface mesh are usually not closed surfaces, the volume of the such segments are obtained by calculating the volume of their convex hulls.

\begin{figure}
    \centering
    \subfigure[]{\includegraphics[width=0.35\textwidth]{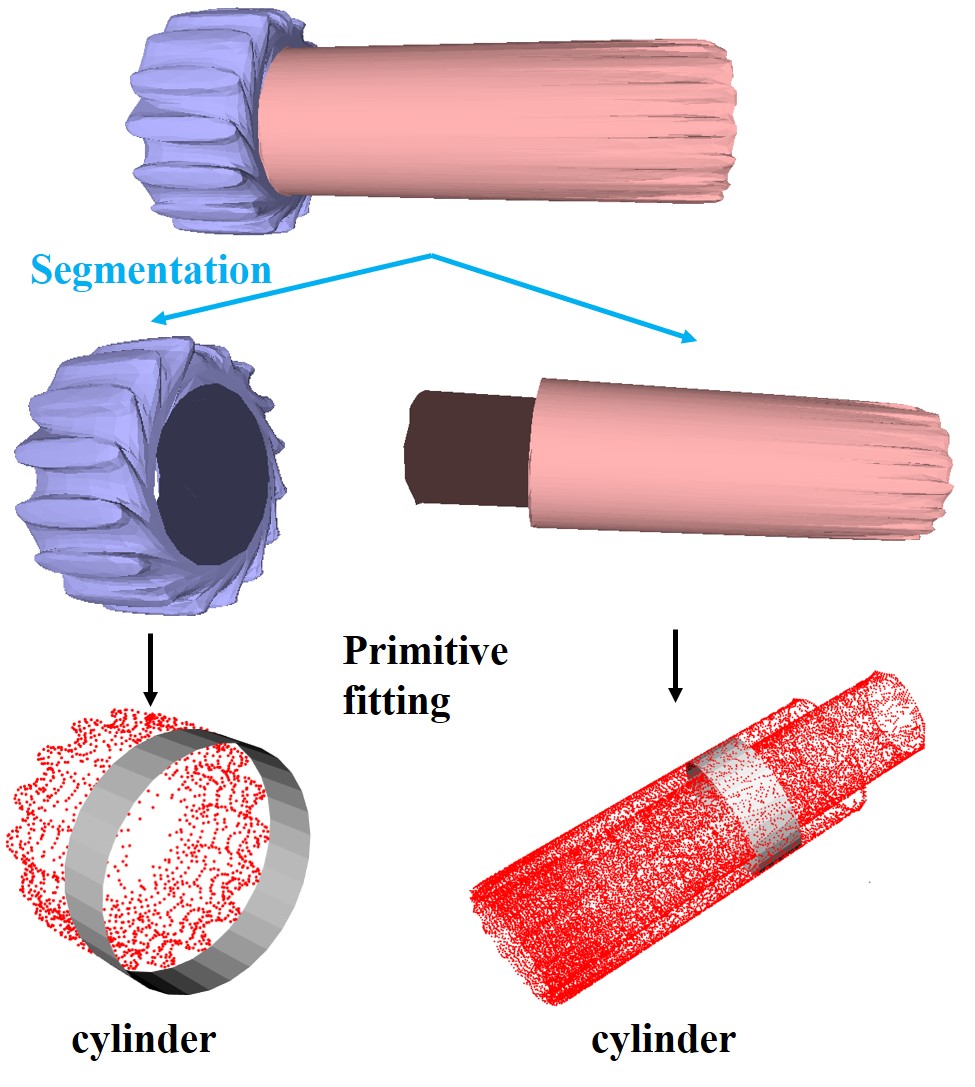}}
    \subfigure[]{\includegraphics[width=0.6\textwidth]{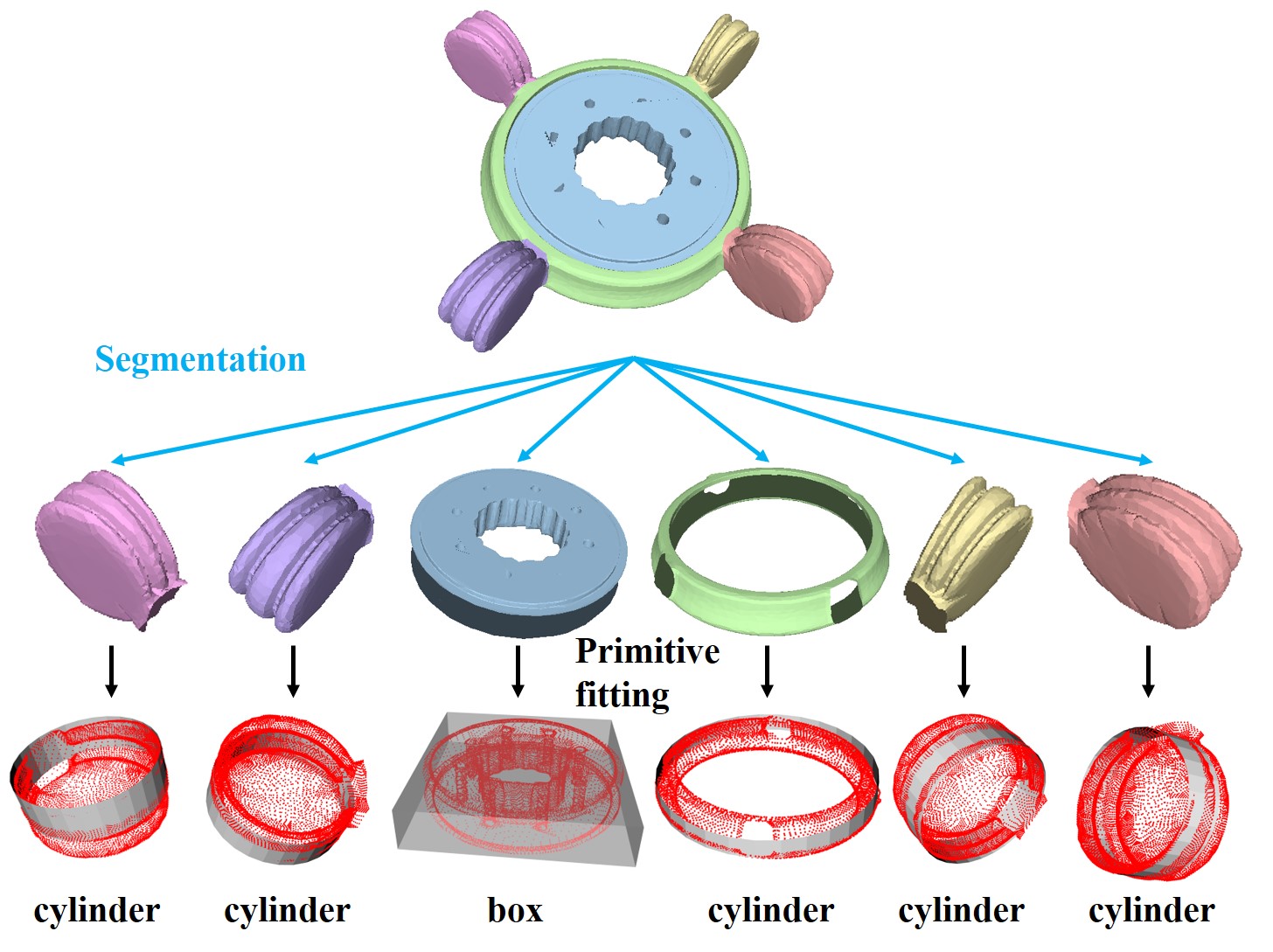}}
    \caption{Two examples of fitting the segmented parts with shape primitives. (a) The inner sleeve shaft is segmented into two parts, both of the two parts are fitted appropriately by cylinders. (b) Five segments of the rotor are more closely fitted by cylinders and the other one is fitted its bounding box, notice that the third segment looks cylindrical but it is empty on the cylindrical surface.}
    \label{fig:primitive_fitting}
\end{figure}

Fig. \ref{fig:primitive_fitting} shows two examples of fitting the segmented parts with primitives. The rotor in Fig. \ref{fig:primitive_fitting} (b) is segmented into 6 parts, we fit all of them with cylinders and bounding boxes, by comparing the volume of the segmented part and the fitted primitives, the appropriate fitting for every segment can be determined. Note that the criteria of determining a better fitting is by comparing the volume, but the precondition is that surface of the fitted primitive is non-empty, such that the gripper can grasp the fitted primitives. And this is why we prefer the RANSAC model fitting provided in PCL \cite{rusu20113d}, it guarantees that the fitted primitive has a non-empty surface for grasping. For example, if we fit the point cloud of the third segment in Fig. \ref{fig:primitive_fitting} (b) with cylinder models, the best fitting we can find may be the cylinder which aligns with its inner hole surface, but the volume difference with the segment is larger comparing to the fitted bounding box, therefore, the better fitting is actually the bounding box. However, there is no box fitting provided in PCL, so we use the bounding box to fit the segment, in this case, we have to check which faces of bounding box are empty and which are not, and then select the non-empty faces to be in contact with the gripper. As a result, five of them can be closely fitted by cylinders and the other one is fitted by its bounding box. The fitted cylinders are represented by gray belts, the height of the cylinder is manually set to 1 cm for visualization, but it can also be calculated from the maximum distance along the cylinder's axis between the points on the segment. Then the corresponding gripper type can be selected for every segment based on the predefined rules. In order to grasp a mechanical component, at least one of its segmented parts should be graspable by the designed grippers, e.g. the gripper for grasping the rotor should be capable of grasping at least one of the 6 segments in Fig. \ref{fig:primitive_fitting} (b).

\subsection{Gripper Parameters}
The maximum and minimum opening widths and finger length are important parameters of the grippers. In order to grasp a segment, the characteristic length of the shape primitive, which is the diameter of the fitted cylinder or the distance between the opposite faces (in touch with the gripper) of its bounding box, must be within the stroke of the gripper. The capability of grasping a segment does not directly impose constraints on the finger length, however, the finger length has to fulfill some requirements in order to satisfy the assembly constraints, for example, the finger should be long enough to avoid collision with the shaft when inserting the shaft sleeve to the shaft. And the constraints on the finger length is described in section 5.2. In addition, the gripper approach direction can be extracted from the fitted primitives. The 3-finger centric gripper should approach the part along the axial direction of the fitted cylinder, and the 2-finger gripper can approach the part as long as the finger surfaces are parallel to the non-empty surfaces of the bounding box.

\section{Evaluation Under Assembly Constraints}
Through mesh segmentation and shape primitive fitting, we have obtained the initial candidate gripper types and parameters for all the segmented parts of the assembly components, however, some of them are not applicable considering the assembly constraints. In this section, we take into account the assembly constraints and finalize the minimum number of grippers for the given assembly task.

\subsection{Assembly Task Specification}
\begin{figure}
    \centering
    \includegraphics[width=0.8\textwidth]{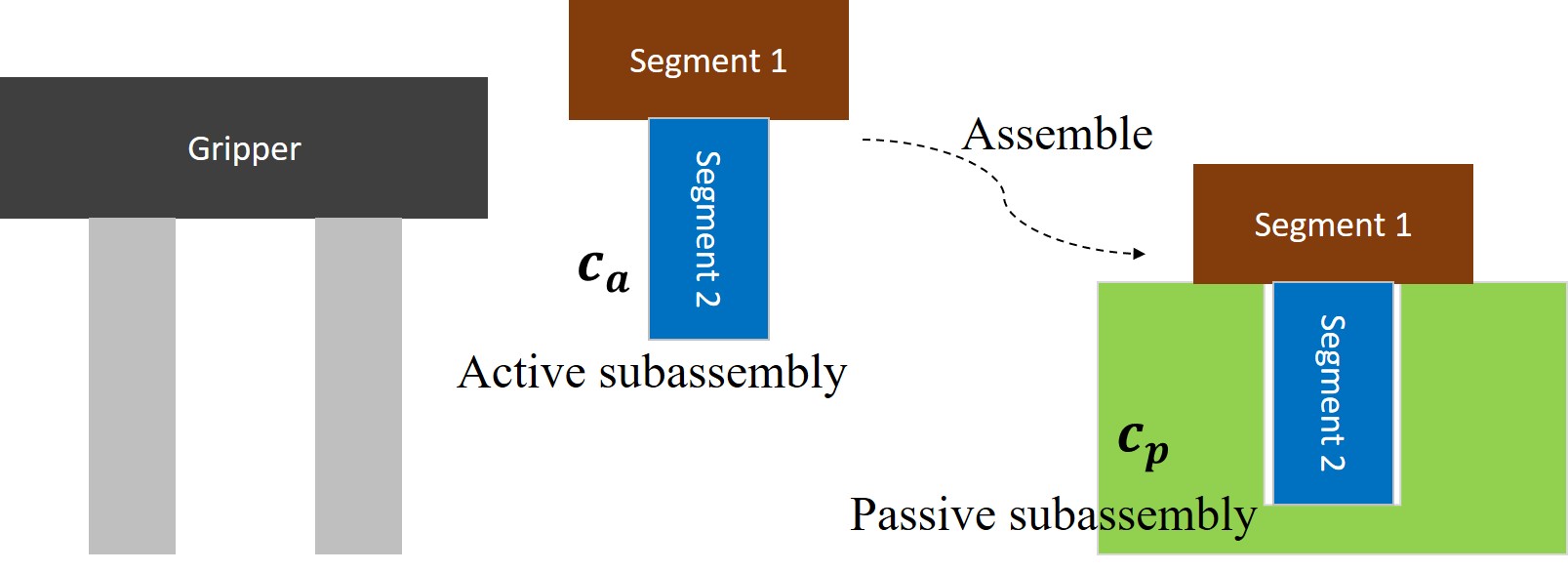}
    \caption{The gripper has to avoid collision with the subassemblies in the assembly task.}
    \label{fig:assembly_constraint}
\end{figure}

Referring to the assembly task decomposition method proposed by Mosemann et al. \cite{mosemann2001automatic}, an assembly task can be represented as a sequence of assembly operations, in each assembly $operation_i$, a new assembly component is added to the existing subassembly. We assume the assembly sequence is already given, then the assembly task is denoted as $Assembly = \{operation_1, operation_2, \ldots,operation_n\}$. Each assembly operation can be represented as $\langle c_a, c_p, ^aT_p, ^{a'}T_p \rangle$, where $c_a$ and $c_p$ are the active and passive subassemblies to be manipulated in the operation, and $^aT_p$ and $^{a'}T_p$ are spatial transformations between active and passive subassemblies before and after the assembly operation, respectively. The active subassembly is the subassembly grasped by the gripper during the assembly operation, and it moves with the gripper until the assembly operation finishes. The passive subassembly is usually fixed in the environment and it serves as an environmental obstacle that should not collide with the gripper. In Fig. \ref{fig:assembly_constraint}, the gripper should grasp the active subassembly $c_a$ to assemble it to the passive subassembly $c_p$.

\subsection{Assembly Constraints}
In an assembly operation $operation_i$, the gripper has to grasp one segment of $c_a$  and change the spatial relationship from $^aT_p$ to $^{a'}T_p$. When grasping $c_a$, not every segment of $c_a$ is suitable for grasping, the affordance of different segments should be taken into account in selecting graspable segments. Moreover, the gripper must avoid the collision with the subassemblies during the assembly.

\begin{figure}
    \centering
    \includegraphics[width=0.6\textwidth]{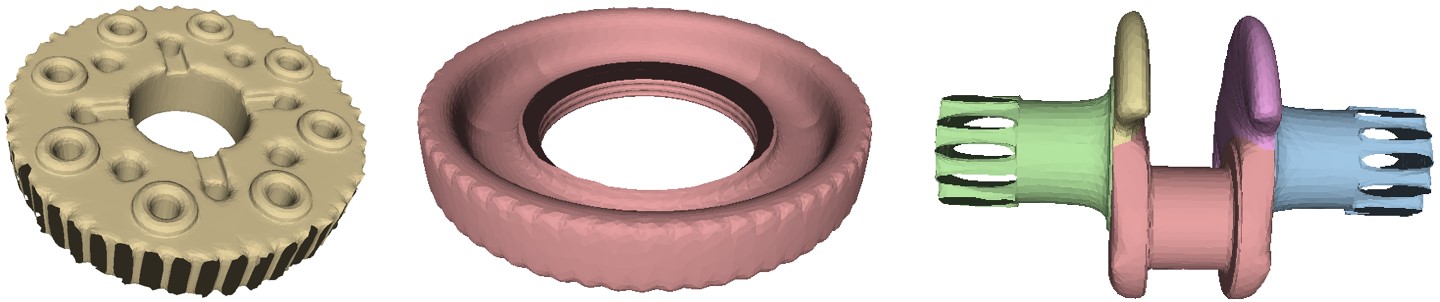}
    \caption{Gear teeth and screw thread do not afford grasping, thus removed from the candidate graspable segments.}
    \label{fig:affordance}
\end{figure}

\subsubsection{Affordance}
Affordance is defined as the possible action on an object or environment \cite{yamanobe2017brief}. In an assembly operation, not all the segments of an assembly component afford grasping. For example, screw thread and gear teeth are mainly used for fastening and transmission, they may be damaged and lose their main affordance if they are directly grasped by the gripper. As illustrated in Fig. \ref{fig:affordance}, some segments are manually removed from the candidate segments for grasping considering their major affordance.

\subsubsection{Collision Avoidance}
The gripper has to avoid collision with the subassemblies during the assembly, the example illustrated by Fig. \ref{fig:assembly_constraint} shows that the gripper will collide with the subassembly if segment 2 is grasped in this assembly operation, thus segment 1 should be selected as the graspable segment. A segment is graspable only if there exists a collision-free grasping pose for the gripper to assemble $c_a$ to $c_p$. To get the graspable segments satisfying the collision avoidance constraint, we plan a set of grasps for each segment and check the collision between the gripper and the subassemblies, the segment is graspable if there is at least one collision-free grasp.

\subsection{Grasp Planning}
\begin{figure}
    \centering
    \subfigure[]{\includegraphics[width=0.25\textwidth]{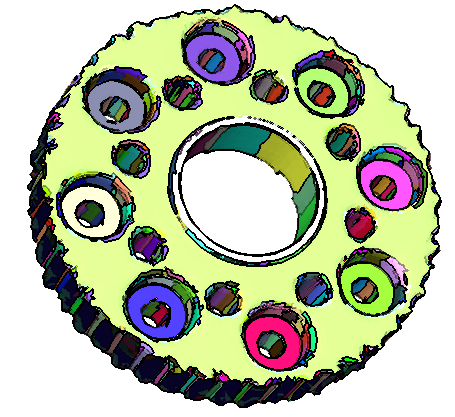}}
    \subfigure[]{\includegraphics[width=0.25\textwidth]{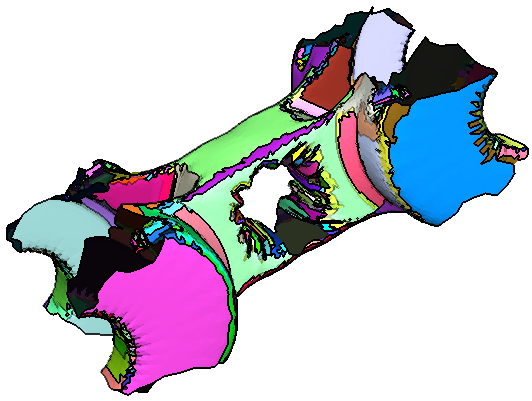}}
    \subfigure[]{\includegraphics[width=0.25\textwidth]{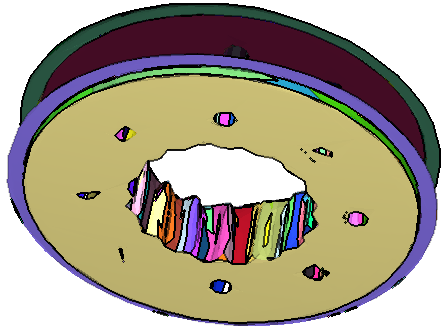}}
    \subfigure[]{\includegraphics[width=0.25\textwidth]{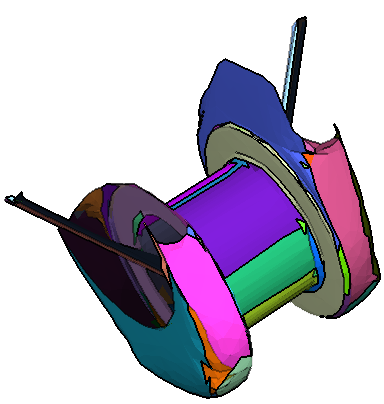}}
    \subfigure[]{\includegraphics[width=0.25\textwidth]{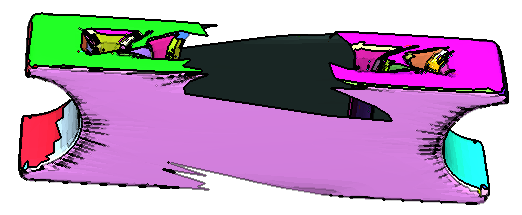}}
    \subfigure[]{\includegraphics[width=0.25\textwidth]{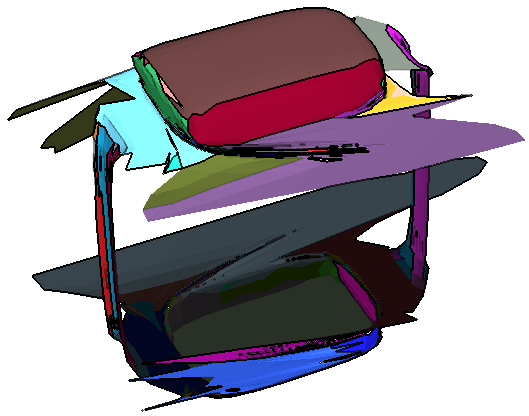}}
    \caption{Planar clustering of the segments to be grasped by 2-finger grippers, different clustered facets are rendered by different colors.}
    \label{fig:planar_clustering}
\end{figure}

\begin{figure}[h]
    \centering
    \subfigure[]{\includegraphics[width=0.2\textwidth]{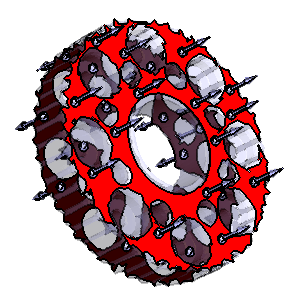}}
    \subfigure[]{\includegraphics[width=0.2\textwidth]{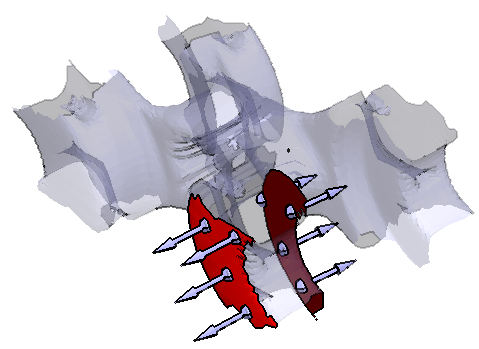}}
    \subfigure[]{\includegraphics[width=0.2\textwidth]{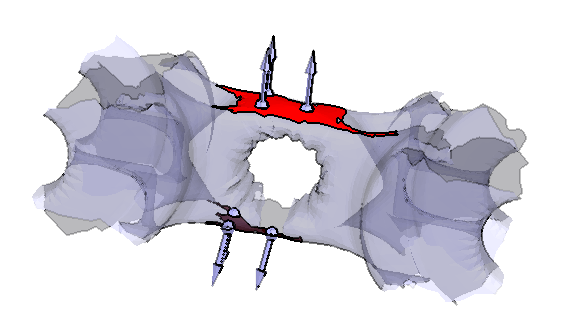}}
    \subfigure[]{\includegraphics[width=0.2\textwidth]{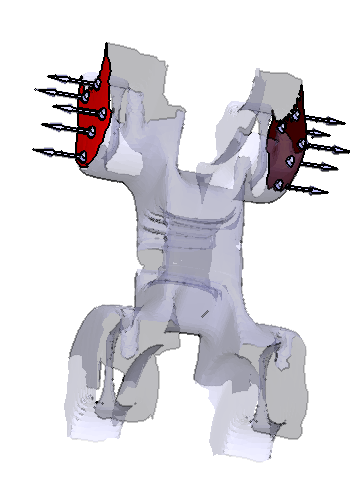}}
    \subfigure[]{\includegraphics[width=0.2\textwidth]{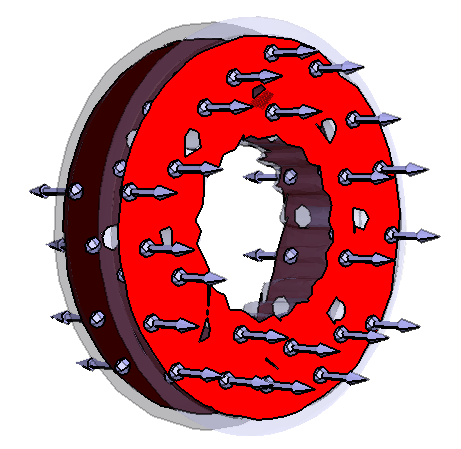}}
    \subfigure[]{\includegraphics[width=0.2\textwidth]{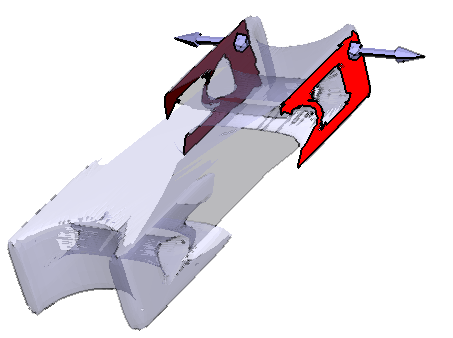}}
    \subfigure[]{\includegraphics[width=0.2\textwidth]{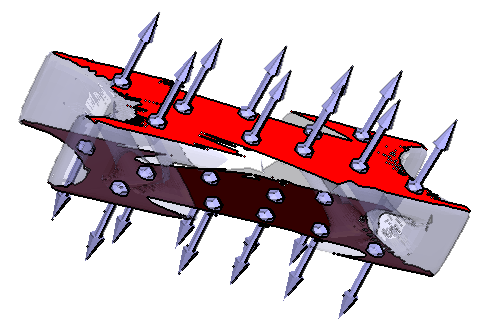}}
    \subfigure[]{\includegraphics[width=0.2\textwidth]{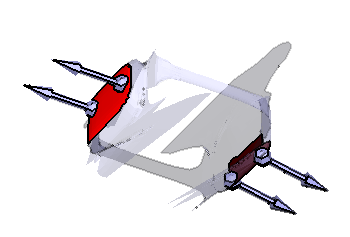}}
    \caption{Pairs of facets and contact points for grasping by 2-finger grippers, the origins of the arrows are the contact points, and the arrows point to the surface normal directions at the contact points.}
    \label{fig:contact_pairs}
\end{figure}

After removing ungraspable segments according to their affordance, grasp planning is performed on the remaining segments to determine if there are collision-free grasps for the segments. For the segments to be grasped by 2-finger parallel grippers, we first use planar clustering \cite{wan2020planning} to cluster the mesh into a set of nearly planar facets, and then search for nearly parallel facets to be in contact the fingers of the gripper, and rotate the gripper around the contact normal to obtain more grasps. Fig. \ref{fig:planar_clustering} shows some examples of the planar clustering, different clustered facets are rendered by different colors. By searching nearly parallel facets from the clustered model, pairs of facets and contact points for grasping are obtained, as shown in Fig. \ref{fig:contact_pairs}. In terms of the segments to be grasped by 3-finger grippers, the grasp can be easily extracted from the fitted cylinder, the axis of the gripper should align with the axis of the cylinder. 

\begin{figure}[h]
    \centering
    \subfigure[]{\includegraphics[width=0.25\textwidth]{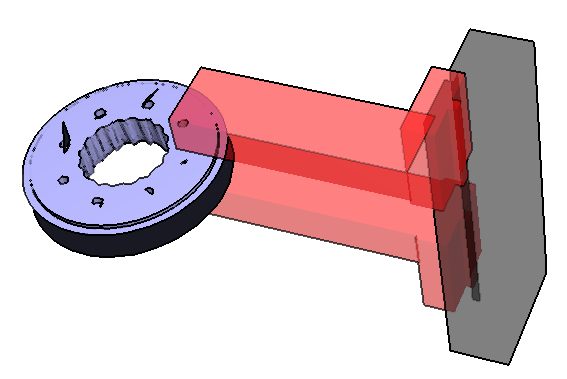}}
    \subfigure[]{\includegraphics[width=0.25\textwidth]{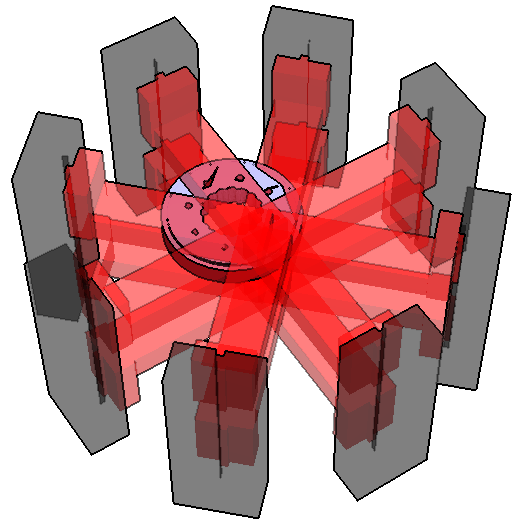}}
    \subfigure[]{\includegraphics[width=0.25\textwidth]{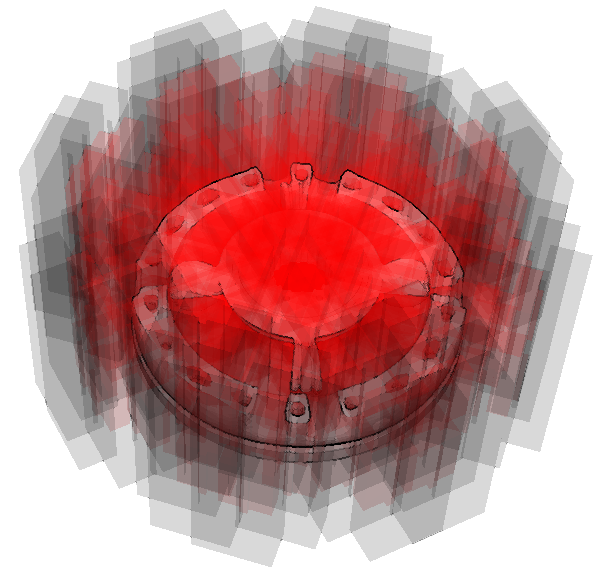}}
    \subfigure[]{\includegraphics[width=0.25\textwidth]{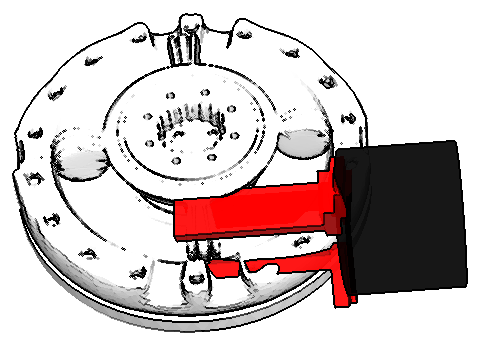}}
    \subfigure[]{\includegraphics[width=0.25\textwidth]{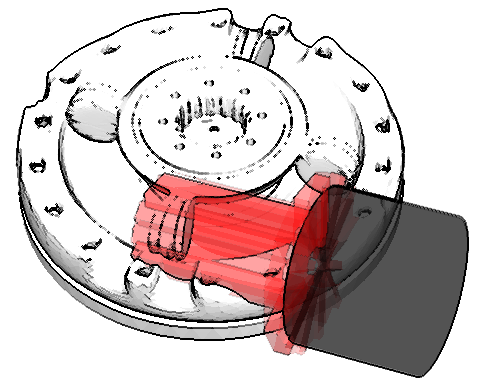}}
    \subfigure[]{\includegraphics[width=0.25\textwidth]{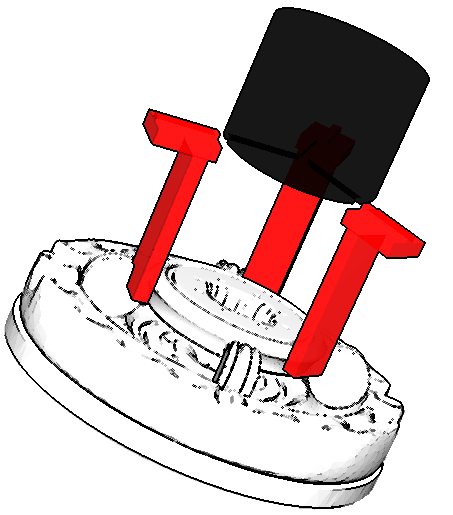}}
    \caption{Mesh segmentation and primitive fitted have determined the gripper types for the segments, then based on the gripper type, grasp planning is performed on the segmented parts to determine if the segment is graspable, by checking if there is at least one collision-free grasp. (a) A planned 2-finger grasp at a pair of contact points. (b) A set of planned 2-finger grasps are obtained by rotating the gripper around the contact normal. (c) Check the collision between the gripper and the subassemblies. (d) A planned 3-finger grasp for the segment. (e) All the planned 3-finger grasps are in collision with the subassemblies. (f) Collision-free grasps can be found for this segment, and it is the only graspable segment for grasping the rotor to perform this assembly operation.}
    \label{fig:collision_check}
\end{figure}

\begin{figure}
    \centering
    \includegraphics[height=\textheight]{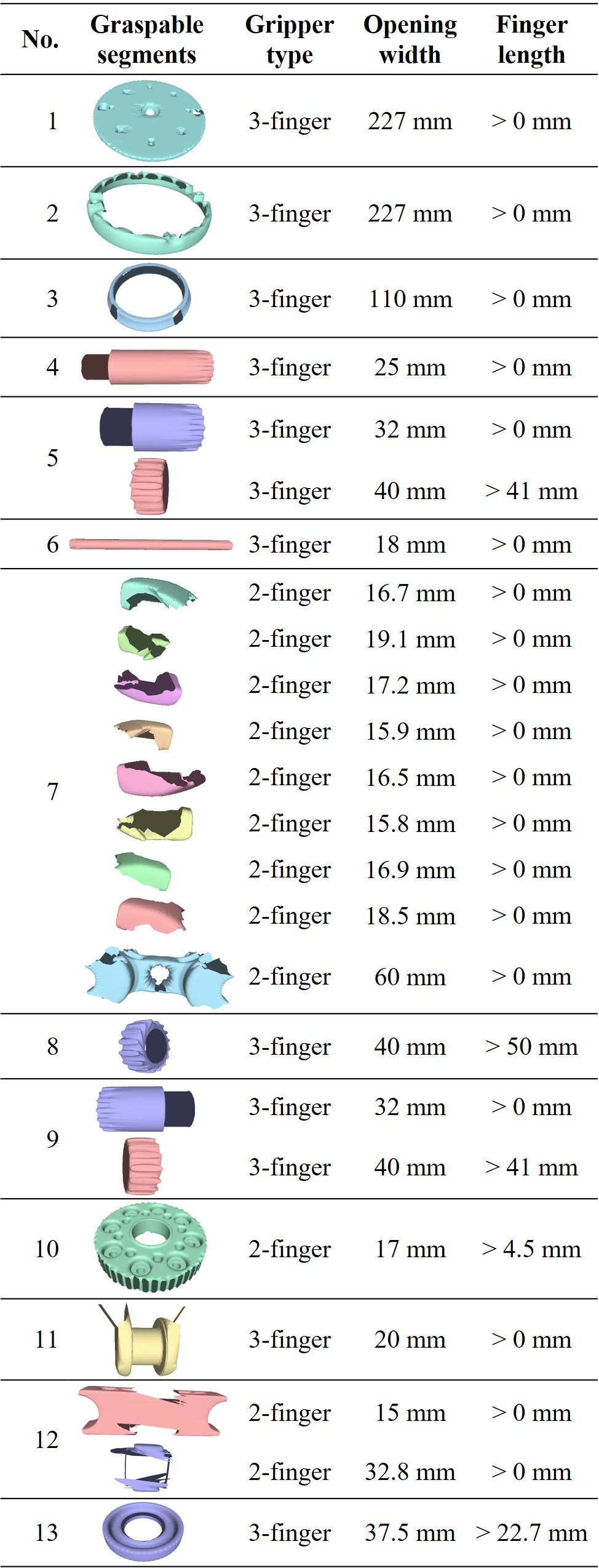}
    \caption{The remaining graspable segments after checking the assembly constraints.}
    \label{fig:all_result}
\end{figure}

The planned grasps are then examined by checking the collision between the gripper and the subassemblies. In Fig. \ref{fig:collision_check}, we explain how grasp planning is used to determine the graspable segment with respect to collision avoidance constraints, and in this example, the sub-task is to assemble the rotor to the subassemblies (including the coolant cover plate and the front cover plate shown in Fig. \ref{fig:all_components_before_processing}). The rotor, as presented in Fig. \ref{fig:primitive_fitting} (b), is comprised of 6 segments. Its third segment should be grasped by a 2-finger parallel jaw gripper, and the planned grasps are shown in Fig. \ref{fig:collision_check} (a), (b) and (c), however, as seen in Fig. \ref{fig:collision_check} (c), all the grasps are in collision with the subassemblies, therefore, the third segment is not graspable. The first, second, fifth and sixth segments of the rotor should be grasped by a 3-finger centric gripper, but they are also not graspable since the planned grasps are in collision with the subassemblies (Fig. \ref{fig:collision_check} (d) \& (e)). The fourth segment of the rotor should be grasped by the 3-finger centric gripper, and it is the only graspable segment, the collision-free grasp is shown in Fig. \ref{fig:collision_check} (f). After the evaluation under the assembly constraints following such process, the remaining graspable segments of every assembly component are listed in Fig. \ref{fig:all_result}, alongside the graspable segments, and there are the constraints on gripper types and parameters for grasping the segments. Among these constraints, the opening width is set to be the characteristic length of the segment, which is the diameter of the fitted cylinder or the distance between the opposite faces (in touch with the gripper) of its bounding box, and the opening width should be within the stroke of the gripper. The finger length should be set such that the finger can contact the segment and also avoid the collision between the gripper and other segments and subassemblies. For example, consider assembling assembly component No. 5 to No. 4 (see Fig. \ref{fig:assembly_experiment} (b1)), if we grasp the second segment of No. 5, then the finger length has to be long enough to avoid colliding with assembly component No. 4 and also the first segment of No. 5, and similarly we obtain the finger length constraint for the second segment of assembly component No. 9 (see Fig. \ref{fig:assembly_experiment} (c2)). For assembly component No. 10, the finger should be longer than the depth of the gear teeth to contact the target segment (see Fig. \ref{fig:assembly_experiment} (c4)). For assembly components No. 8 and No. 13, the finger length should be above the threshold to avoid colliding with the shaft (No. 6) (see Fig. \ref{fig:assembly_experiment} (d1) and (d5)). 

\subsection{Minimize the Number of Grippers}
Some of the assembly components can be commonly grasped by the same gripper, thus the total cost of grippers can be cut down by reducing the number of grippers for the assembly task. From the previous analysis, we have obtained the graspable segments from all the assembly components, and every assembly component $c_i$ imposes a set of constraints on the gripper, such as the number of fingers $F_i$, opening width $W_i$, minimum finger length $L_i^-$ and maximum finger length $L_i^+$. We denote the constraints for assembly component $c_i$ as $\mathcal{C}_i = \{F_i, W_i, L_i^-, L_i^+\}$, $i=1,2,\dots,M$, $M$ is the number of assembly components, if an assembly component $c_i$ has $m$ graspable segments $\{c_{i1}, c_{i2},\dots,c_{im}\}$, then $\mathcal{C}_i= \mathcal{C}_{i1} \cup \mathcal{C}_{i2} \cup \dots \cup \mathcal{C}_{im}$, where $\mathcal{C}_{ij}$ is the constraints imposed by segment $c_{ij}$ of $c_i$. We generate $N$ gripper parameters $\{P_1, P_2, \dots, P_N\}, P_j = \{F_j, W_j^-, W_j^+, L_j\}$, covering the minimum and maximmum gripper parameters, then determine the minimal subset out of the $N$ gripper parameters that can grasp all the assembly components, and the problem is formulated as follows,
\begin{equation}
\begin{array}{ll@{}ll}
\text{minimize} & \sum_{j=1}^N x_j \\
\text{subject to}
 & x_j = \{0, 1\} \\
 & \sum_{j=1}^N a_{1j}x_j \geq 1 \\
 & \sum_{j=1}^N a_{2j}x_j \geq 1 \\
 & \vdots \\
 & \sum_{j=1}^N a_{Mj}x_j \geq 1
\end{array}
\end{equation}
\begin{equation}
\left(
\begin{aligned}
F_j = F_i \\ 
L_i^-<L_j<L_i^+ \\ 
W_j^-<W_i<W_j^+
\end{aligned}
\right)
\end{equation}
where $i \in \{1,2,\dots,M\}, j \in \{1,2,\dots,N\}$, $M$ is the number of assembly components, $N$ is the number of gripper parameter samples, $a_{ij}$ is 1 if Eq. (2) is satisfied, otherwise $a_{ij} = 0$, $W_j^-$ and $W_j^+$ are the minimum and maximum opening widths of a gripper, respectively, $L_j$ is the finger length, and $x_j$ is 1 if $j$-th gripper parameter sample $P_j$ is used and is 0 otherwise. Eq. (2) lists the conditions for the the gripper with parameter $P_j$ to grasp the assembly component $c_i$, and the inequalities in Eq. (1) ensure that every assembly component $c_i$ can be grasped by at least one gripper.

This is a multidimensional 0–1 knapsack problem \cite{freville2004multidimensional}, to solve it, we have to, (1) obtain $N$ gripper parameters and then determine a minimal subset out of $N$ subject to these constraints, and (2) figure out all the coefficients $a_{ij}$. For generating $N$ samples of gripper parameters, we separate the samples into two groups, the first group is for 2-finger grippers ($F_j=2$), and the second group is for 3-finger grippers ($F_j=3$). For each group, we set the upper and lower bounds of the maximum opening width $W_{upper}^+, W_{lower}^+$ to be just enough to cover the maximum and minimum values of opening width $W_i$, that is $W_{upper}^+ = \{W_i\}_{max}, W_{lower}^+ - d = \{W_i\}_{min}$, where $d$ is the stroke of the gripper, if $\{W_i\}_{max} - \{W_i\}_{min} \leq d$, then we set $W_{upper}^+ = W_{lower}^+ = \{W_i\}_{max}$. Notice that, we do not assume the gripper can be fully closed, instead we assume the gripper has a certain stroke $d$, or more specifically, $d_{2finger}$ for the 2-finger gripper and $d_{3finger}$ for the 3-finger gripper, following such assumption, the minimum opening width can be directly derived as: $W_j^- = W_j^+ - d$, and the strokes of the grippers that we use are introduced in Section 6. The upper and lower bounds of finger length $L_{upper}, L_{lower}$ are set to be the maximum and minimum finger lengths among these constraints. So far, we obtain the bounds for the opening width and the finger length for each group, then for each group we uniformly generate $n$ values of $W_j^+$ from range $[W_{lower}^+, W_{upper}^+]$ and $m$ values of $L_j$ from range $[L_{lower}, L_{upper}]$. Specifically, we obtain $n_1$ values of $W_j^+$ and $m_1$ values of $L_j$ for the first group, and $n_2$ values of $W_j^+$ and $m_2$ values of $L_j$ for the second group. Therefore, we totally have $N = n_1 \times m_1 + n_2 \times m_2$ sets of gripper parameters for the two parameter groups, and the coefficients $a_{ij}$ is obtained by checking if Eq. (2) is satisfied for assembly constraint $\mathcal{C}_i$ and gripper parameter $P_j$, then this 0-1 knapsack problem can be solved by using branch-and-bound method \cite{kolesar1967branch}. Since there is no upper bound for the finger length in our case, the upper bound can be flexibly set to a value as long as it is large than the maximum lower limit of these finger length constraints, the obtained solutions for our case is presented in Section 6.

\subsection{Discussion and Limitation}
In this research, we assume that the target assembly components can be well decomposed into boxes and cylinders, and we only use two types of grippers, which are 2-finger parallel jaw grippers and 3-finger centric grippers. To cope with more complex shapes, we have to use more shape primitives, such as cone and pyramid. In addition to affordance and collision avoidance described above, there are other aspects to be considered for further improvement.

\subsubsection{Stability of Grasping Different Segments}
In an assembly operation, grasping different segments may result in different force/torque distribution. Consider assembling the carrier to the shaft (Fig. \ref{fig:all_components_before_processing}), if the grasping contact positions are not symmetric about the shaft, Fig. \ref{fig:assembly_experiment} (b5) shows an example of such situation, it will lead to uneven normal force between shaft and hole, which may result in insertion failure, or even damage the assembly components. Therefore, it is necessary to analyze the contact force distribution when selecting suitable segment for grasping during the assembly.

\subsubsection{Finer Finger Design}
The assembly components must be stably grasped without slipping during the assembly, in which the external forces include gravity, assembly force, etc. It is necessary to fine-tune the shape of the fingertip surface to increase the contact area with the object, especially when the object surface is curved. Assuming the soft-finger contact model, we can calculate the contact area from the relative curvature between fingertip surface and object surface \cite{harada2014stability}, then an appropriate fingertip surface curvature that ensures the grasp stability can be determined. Fig. \ref{fig:gripper_rules} (a) illustrates the situation of the maximum torque caused by gravity, and it should be balanced by the torsional friction exerted by the soft finger contact.

\section{Experiment}
\begin{figure}
    \centering
    \includegraphics[width=0.6\textwidth]{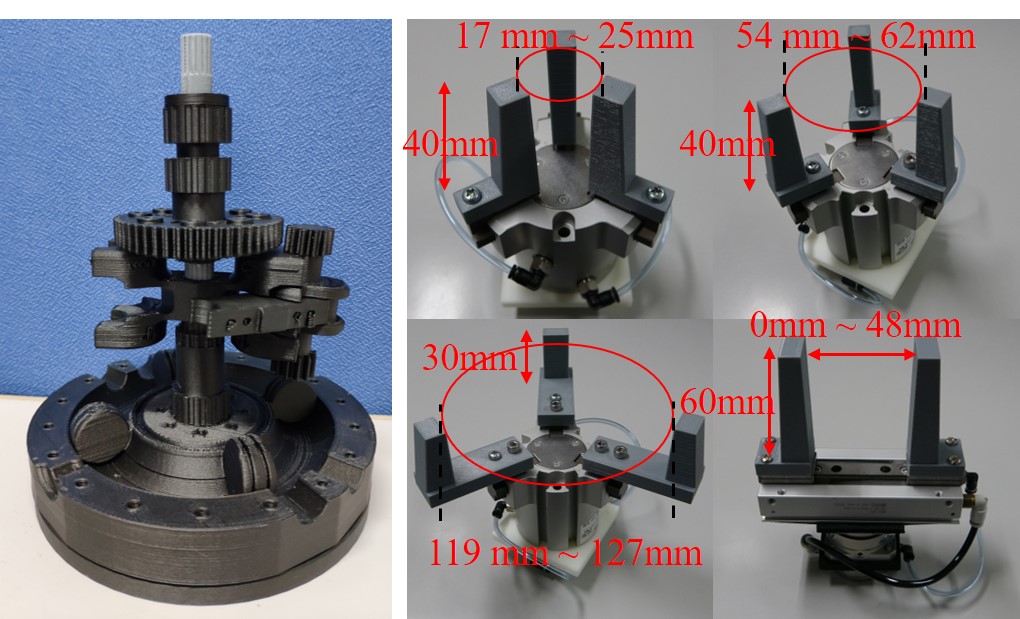}
    \caption{(Left): The product to be assembled. (Right): The designed 4 grippers in their maximum opening state, the strokes are 8mm for 3-finger air chuck and 48mm for the 2-finger air chuck, respectively.}
    \label{fig:product_grippers}
\end{figure}

In this section, the effectiveness and feasibility of our approach are verified by assembling a part of an industrial product using the designed grippers. Considering the limit of our 3D printer, the product is scaled to 55\% of its original size and printed out as shown in Fig. \ref{fig:product_grippers} (Left). The grippers are constructed by attaching printed fingers to air chucks, the 2-finger gripper is constructed by attaching 2 fingers on a SMC MHF2-12D2 air chuck (stroke: 48mm, 0mm to 48 mm), the 3-finger gripper is constructed by attaching 3 fingers on a SMC MHSL3-32D air chuck (stroke: 8mm, 34 mm to 42 mm), the stroke of an air chuck determines the difference between the maximum and minimum opening widths $(W_j^+ - W_j^-)$. According to the strokes of air chucks we use and the scaled dimensions of the assembly constraints, the feasible solutions for the gripper parameters are: $\{2, 0, 33, 30\}, \{3, 14, 22, 30\}, \{3, 52.5, 60.5, 30\}, \{3, 116.9, 124.9, 30\}$, that is one 2-finger gripper with a maximum opening width of 33 mm, and three 3-finger grippers with maximum opening widths of 22 mm, 60.5 mm and 124.9 mm are required. We model and print out the fingers and attach them to the air chucks, and we calculate the position where the fingers should be fixed on the air chucks such that the maximum and minimum opening widths correspond to the solutions. For the 2-finger gripper, since the stroke is 48 mm, which is larger than the difference between the maximum and minimum opening widths in solution $\{2, 0, 33, 30\}$, so we simply design the 2-finger gripper with the opening width ranging from 0 mm to 48 mm, which covers the range of opening width in the solution and works equally well. The actual maximum opening widths for the other three 3-finger grippers are slightly larger than the calculation results to account for the uncertainties. Besides, there is no upper bound on finger length, it is free to set finger length above 30 mm, and all the gripper parameters of the actually gripper used in our experiment are shown in Fig. \ref{fig:product_grippers} (Right).

\begin{figure*}
    \centering
    \subfigure{\includegraphics[width=0.07\linewidth]{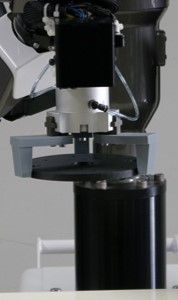}}
    \subfigure{\includegraphics[width=0.07\linewidth]{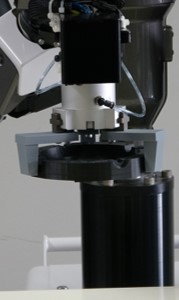}}
    \subfigure{\includegraphics[width=0.07\linewidth]{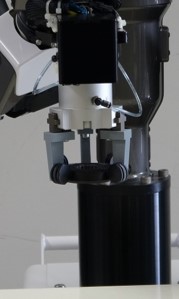}}
    \subfigure{\includegraphics[width=0.07\linewidth]{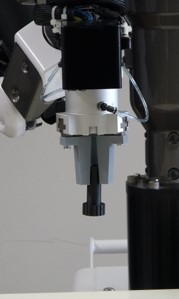}}
    \subfigure{\includegraphics[width=0.07\linewidth]{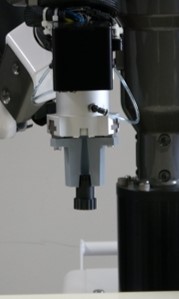}}
    \subfigure{\includegraphics[width=0.07\linewidth]{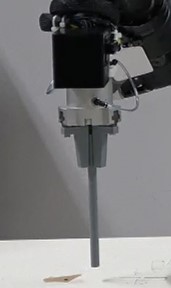}}
    \subfigure{\includegraphics[width=0.07\linewidth]{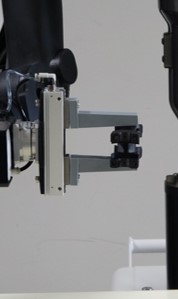}}
    \subfigure{\includegraphics[width=0.07\linewidth]{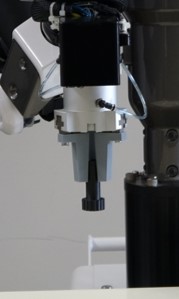}}
    \subfigure{\includegraphics[width=0.07\linewidth]{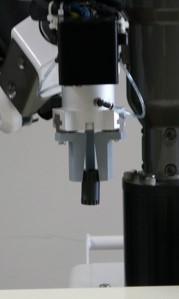}}
    \subfigure{\includegraphics[width=0.07\linewidth]{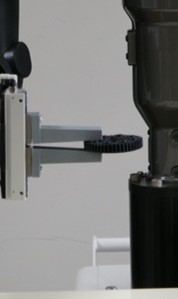}}
    \subfigure{\includegraphics[width=0.07\linewidth]{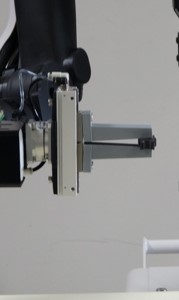}}
    \subfigure{\includegraphics[width=0.07\linewidth]{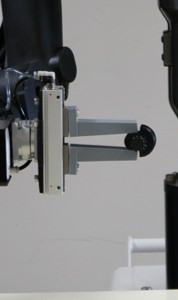}}
    \subfigure{\includegraphics[width=0.07\linewidth]{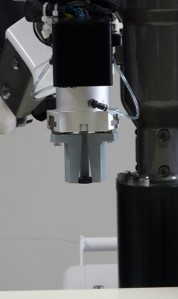}}
    \caption{Designed 4 grippers are able to firmly grasp all the 13 assembly components.}
    \label{fig:grasping_all_parts}
\end{figure*}

\begin{figure*}
    \centering
    \includegraphics[width=\textwidth]{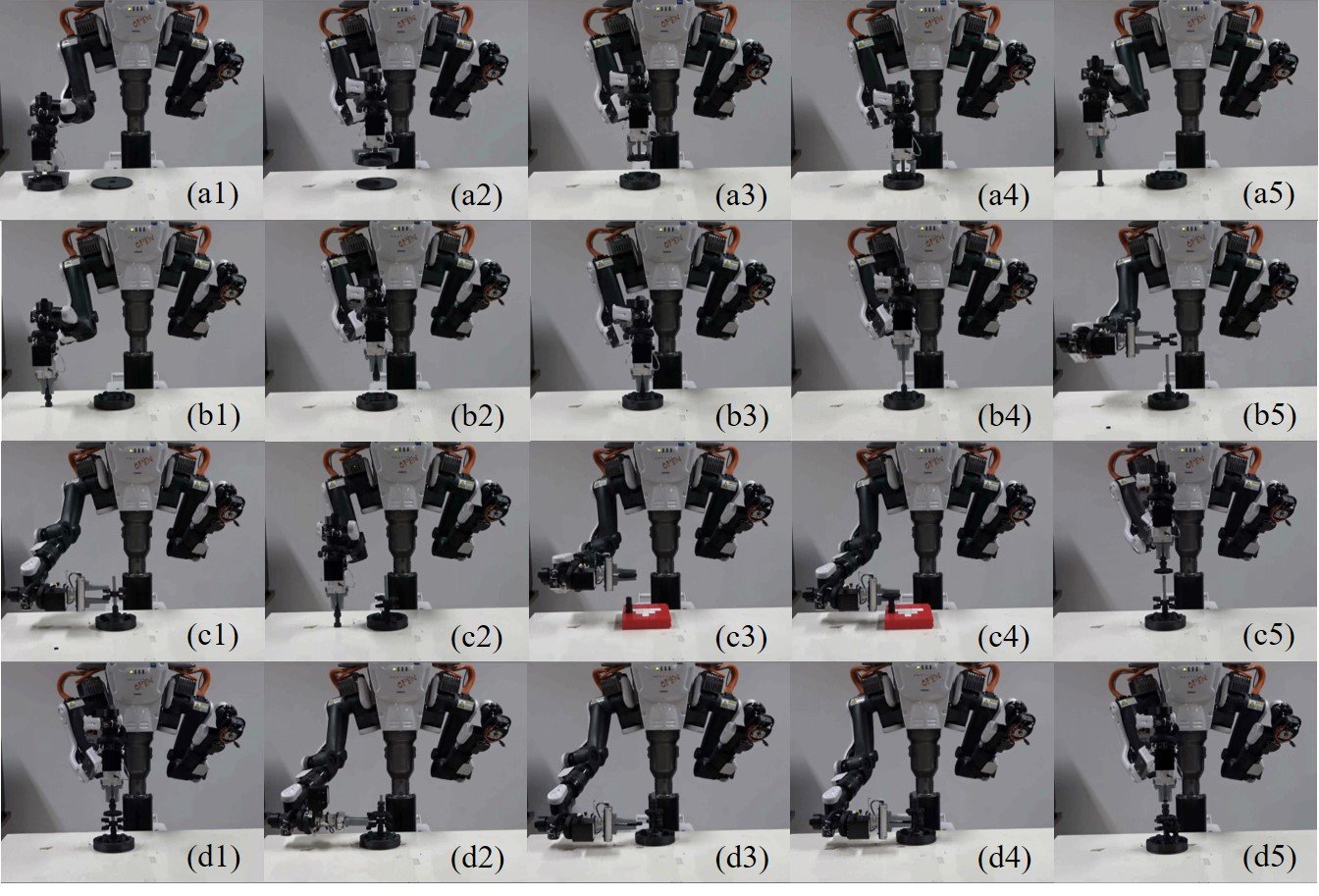}
    \caption{The robot can successfully assemble the product, and there is no collision between the gripper and the subassembly, the red object appearing in (c3) and (c4) is used to support the subassembly.}
    \label{fig:assembly_experiment}
\end{figure*}

We performed the assembly experiment on a NEXTAGE robot from Kawada Robotics Inc., as shown in Fig. \ref{fig:grasping_all_parts}, all the 13 assembly components can be firmly grasped by using the designed 4 grippers. We assume the assembly sequence is known, the target segment of an assembly component for grasping can be obtained from the previous analysis, then the robot is able to successfully complete the task without collision with the subassemblies during the assembly, as shown in Fig. \ref{fig:assembly_experiment}. 

\section{Conclusions and Future Work}
Tackling the challenges of designing grippers for an assembly task, we presented a structured approach to selecting and designing the grippers. The input for our approach are the assembly specification and the geometrical models of the assembly components. In the first phase, the assembly components with complex shapes are segmented into simpler parts, then the segmented parts are fitted with shape primitives. By defining the correspondence between simple shape primitives and gripper types, suitable gripper types and parameters can be determined from the results of mesh segmentation and primitive fitting. In the second phase, the results in the first phase are examined under the assembly constraints, afterwards, the number of grippers is optimized by finding a minimal set of gripper parameters that satisfy the constraints imposed by all the assembly components. Finally, the effectiveness of designed grippers is confirmed by the assembly experiment.

In the future, the current work can be improved from several aspects: (1) We consider exploring more powerful mesh segmentation methods \cite{le2017primitive} to decompose the assembly components, the affordance of the assembly component will be taken into account in the segmentation. (2) More shape primitives such as cone and pyramid can be used to improve the ability of fitting more complex shapes. (3) The representation of the assembly task and constraints can be refined, and classifying the basic assembly operations (such as peg-in-hole) can further automate the design process. (4) The shape of the fingertip surface can be fine-tuned to increase the contact area, thus the grasp stability is improved.

\bibliographystyle{ieeetr}
\bibliography{ref.bib}

\label{lastpage}

\end{document}